%% file: main.tex
\documentclass[10pt,twocolumn,letterpaper]{article}
\usepackage[T1]{fontenc}
\usepackage[utf8]{inputenc}
\usepackage{times}
\usepackage{microtype}
\usepackage{amsmath,amssymb,bm}
\input{math_commands.tex}

\usepackage{etoolbox}
\usepackage[dvipsnames]{xcolor}
\usepackage{graphicx,booktabs,array,multirow,tabularx,longtable}
\usepackage[format=plain,labelformat=simple,labelsep=period,font=small,compatibility=false]{caption}
\usepackage[font=footnotesize,skip=3pt,subrefformat=parens]{subcaption}
\usepackage{stfloats,placeins}
\usepackage[shortlabels,inline]{enumitem}
\setlist[itemize]{noitemsep,leftmargin=*,topsep=0pt}
\setlist[enumerate]{noitemsep,leftmargin=*,topsep=0pt}
\usepackage{wrapfig,capt-of,needspace,verbatim}
\usepackage[numbers,sort&compress]{natbib}
\usepackage[hyphens]{url}
\definecolor{linkblue}{rgb}{0.12,0.33,0.57}
\usepackage[breaklinks,colorlinks,allcolors=linkblue]{hyperref}
\hypersetup{
  pdftitle={World2HOSI: A Multi-Expert Harness for Long-Horizon Human-Object-Scene Interaction Generation},
  pdfauthor={Jiushu Li, Haoran Wang, Yumeng Liu, Zhiyang Dou, Yuan Liu, Yuexin Ma, Cheng Lin}
}
\renewcommand{\paragraph}[1]{\par\addvspace{.5em}\noindent\textbf{#1}\ }

\title{ World2HOSI: A Multi-Expert Harness for\\
Long-Horizon Human-Object-Scene Interaction Generation}
\author{%
  \textbf{Jiushu Li}\textsuperscript{1}\hspace{0.78em}%
  \textbf{Haoran Wang}\textsuperscript{1}\hspace{0.78em}%
  \textbf{Yumeng Liu}\textsuperscript{2}\hspace{0.78em}%
  \textbf{Zhiyang Dou}\textsuperscript{3}\hspace{0.78em}%
  \textbf{Yuan Liu}\textsuperscript{4}\hspace{0.78em}%
  \textbf{Yuexin Ma}\textsuperscript{5}\hspace{0.78em}%
  \textbf{Cheng Lin}\textsuperscript{1,\ensuremath{\dagger}}\\[3pt]
  \textsuperscript{1}Macau University of Science and Technology\hspace{1.75em}%
  \textsuperscript{2}University of Science and Technology of China\\
  \textsuperscript{3}Massachusetts Institute of Technology\hspace{1.75em}%
  \textsuperscript{4}The Hong Kong University of Science and Technology\\
  \textsuperscript{5}ShanghaiTech University}
\date{}

\makeatletter
\font\paper@subheading=ptmb scaled 1100
\font\paper@minorheading=ptmb scaled 1000
\newcommand{\paper@section}{\@startsection{section}{1}{\z@}{-10pt plus -2pt minus -2pt}{7pt}{\large\bfseries}}
\newcommand{\paper@subsection}{\@startsection{subsection}{2}{\z@}{-8pt plus -2pt minus -2pt}{5pt}{\paper@subheading}}
\newcommand{\paper@subsubsection}{\@startsection{subsubsection}{3}{\z@}{-6pt plus -2pt minus -2pt}{3pt}{\paper@minorheading}}
\newcommand{\paper@heading}[1]{\@ifstar{#1*}{\paper@numbered{#1}}}
\newcommand{\paper@numbered}[2]{#1{\texorpdfstring{\hskip -1em.~}{}#2}}
\renewcommand{\section}{\paper@heading\paper@section}
\renewcommand{\subsection}{\paper@heading\paper@subsection}
\renewcommand{\subsubsection}{\paper@heading\paper@subsubsection}
\renewenvironment{abstract}{%
   \centerline{\large\bfseries Abstract}%
   \vspace*{10pt}\noindent\itshape\ignorespaces
}{\vspace*{10pt}}
\renewcommand{\@maketitle}{%
  \newpage\null\vskip 0pt
  \begin{center}
    {\Large\bfseries\@title\par}
    \vspace{6pt}
    {\normalsize\lineskip .5em\begin{tabular}[t]{c}\@author\end{tabular}\par}
    \vspace{8pt}
  \end{center}
  \input{figures/fig01_teaser}}
\makeatother

\begin{document}
\raggedbottom
\maketitle
\begingroup
\renewcommand{\thefootnote}{\fnsymbol{footnote}}
\footnotetext[2]{Corresponding author.}
\endgroup

\begin{abstract}
\input{sections/00_abstract}

\end{abstract}
\input{sections/01_introduction}
\input{sections/02_related_work}
\input{sections/03_method}

\input{sections/04_experiments}

\input{sections/05_conclusion}

\clearpage
\section*{LLM Usage Statement}
\addcontentsline{toc}{section}{LLM Usage Statement}
\input{sections/06_ai_usage}

{\small
\bibliographystyle{ieeenat_fullname}
\bibliography{references}
}

\clearpage
\setlength{\textheight}{8.875in}
\setcounter{dbltopnumber}{1}
\setlength{\textfloatsep}{20pt plus 2pt minus 4pt}
\setlength{\dbltextfloatsep}{20pt plus 2pt minus 4pt}
\setlength{\floatsep}{12pt plus 2pt minus 2pt}
\setlength{\dblfloatsep}{12pt plus 2pt minus 2pt}
\setlength{\abovecaptionskip}{10pt}
\twocolumn[{%
  \centering
  \Large
  \textbf{World2HOSI: A Multi-Expert Harness for\\
  Long-Horizon Human-Object-Scene Interaction Generation}\\
  \vspace{0.5em}Supplementary Material\\
  \vspace{1.0em}
}]
\appendix
\input{sections/appendix}
\end{document}

%% file: math_commands.tex
\usepackage{amsmath,amsfonts,bm}

\def\eqref#1{equation~\ref{#1}}
\def\1{\bm{1}}

\DeclareMathAlphabet{\mathsfit}{\encodingdefault}{\sfdefault}{m}{sl}
\SetMathAlphabet{\mathsfit}{bold}{\encodingdefault}{\sfdefault}{bx}{n}

%% file: figures/fig01_teaser.tex
\begin{minipage}{\textwidth}
\centering
\includegraphics[trim=0 0 0 65bp,clip,width=\linewidth]{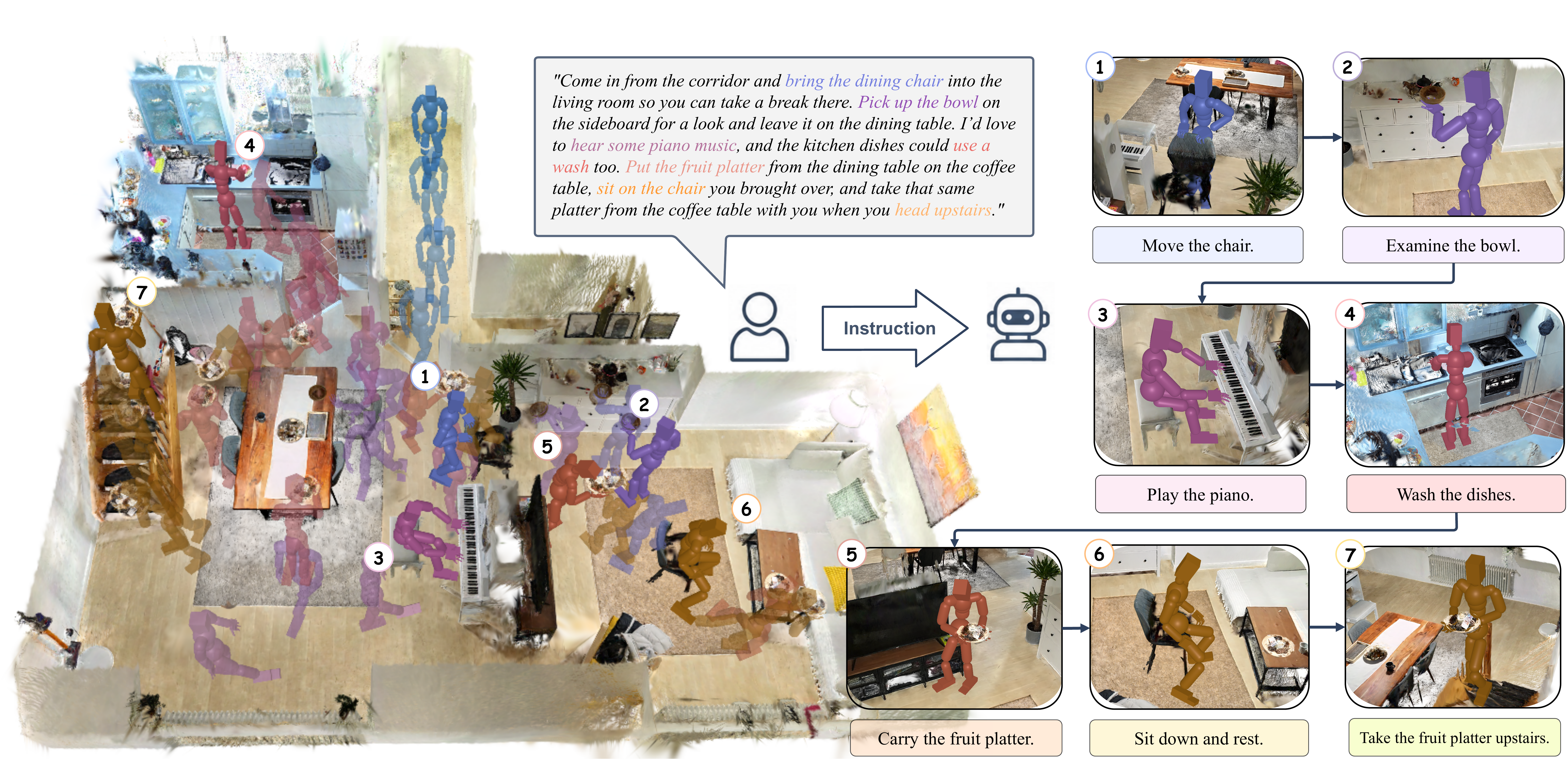}
\captionof{figure}{Long-horizon HOSI generation in reconstructed 3D worlds. World2HOSI coordinates motion experts from text instructions to produce navigation, object manipulation, and scene interaction (1–7) in an updated world.}
\label{fig:teaser}
\end{minipage}\par\medskip

%% file: sections/00_abstract.tex
\looseness=-1 Long-horizon human–object–scene interaction (HOSI) is central to embodied AI, animation, and game development. Existing human-object interaction (HOI) and human–scene interaction (HSI) models excel in their respective domains, but none alone covers tasks requiring multiple interaction skills. Training unified HOSI models is hindered by scarce and costly data. We introduce World2HOSI, a multi-expert orchestration harness that generates continuous HOSI data from 3D Gaussian Splatting (3DGS) scenes and natural language instructions. Our key insight is that long-horizon HOSI hinges not on a stronger single model but on orchestration. Given a shared, structured scene state and an orchestration harness grounded in it, existing experts can jointly accomplish tasks beyond the reach of any of them alone. World2HOSI realizes this insight as a closed loop around a persistent scene state built from decoupled background and object assets. A harness plans action stages over this state, invokes experts via their native conditioning interfaces, and writes each action's effects back, so that every invocation is grounded in the evolving world. To reduce physical artifacts, bounded residual optimization and contact-driven physics simulation further ensure the physical plausibility of human motion and object manipulation, respectively, while preserving interaction intent. Experiments and a perceptual user study show that World2HOSI achieves the best task completion and interaction quality on the evaluated HOSI tasks, while generating more diverse data. Project Page: \href{https://nintreesssss.github.io/World2HOSI}{nintreesssss.github.io/World2HOSI}.

%% file: sections/01_introduction.tex
\section{Introduction}
\label{sec:introduction}

\looseness=-1 Human-centric 3D environments provide a foundation for embodied AI training, animation, and game development. Beyond high-fidelity reconstruction, these applications require natural interactions between humans, objects, and scenes, termed human-object-scene interaction (HOSI). Motion data capturing these interactions support training and evaluation in computer graphics, embodied AI, and agent learning, but remain scarce. GRAB~\citep{GRAB:2020}, BEHAVE~\citep{bhatnagar22behave}, and TRUMANS~\citep{jiang2024scaling} capture human motion together with objects or scenes to support interaction learning. Yet costly motion capture struggles to cover increasingly diverse indoor layouts, object combinations, and multi-stage tasks. This motivates using existing scene and motion resources to produce richer HOSI data.

\looseness=-1 Generative models offer an alternative route to data synthesis. BEDLAM~\citep{Black_CVPR_2023} shows that networks trained entirely on synthetic data can achieve state-of-the-art human pose and shape estimation. InHabit~\citep{kister2026inhabit} leverages interaction priors in image foundation models to synthesize 3D human poses in scenes, benefiting downstream tasks such as contact estimation. For 4D interactions, existing models offer complementary capabilities: human-scene interaction (HSI) models condition motion on static furniture geometry, while human-object interaction (HOI) models coordinate humans and dynamic objects~\citep{Kimodo2026,zhao2023synthesizing,hoifhli}. However, individual models lack the full chain of capabilities required for long-horizon HOSI. For example, ``move the chair and then sit down'' combines dynamic manipulation with static interaction, and requires subsequent motion to reflect the chair's new location. Either capability alone is insufficient. A unified HOSI generator can broaden this capability coverage, but doing so requires training data spanning diverse combinations of actions, objects, and scenes, which remain scarce. InfBaGel's hybrid-data training indirectly reflects this scarcity~\citep{zou2026infbagel}. Training an additional unified generator for data production would reintroduce a dependence on the very data that remain scarce. Orchestrating complementary experts through conditioning interfaces is therefore essential to expanding task coverage without additional motion-generator training.

\looseness=-1 Therefore, our goal is to generate text-guided, long-horizon 4D interactions in realistic reconstructed scenes by coordinating motion experts. We build on 3D Gaussian Splatting (3DGS), which combines rich appearance with efficient novel-view rendering to provide realistic worlds for data synthesis~\citep{kerbl3Dgaussians}. However, the central challenge lies not merely in synthesizing photorealistic scenes or concatenating expert outputs, but in enabling the underlying scene representation to dynamically update spatial and physical constraints for each expert as prior interactions change the environment. A scheduler is required to connect task intent to a shared scene state that persists across expert invocations.

\looseness=-1 Realizing this goal presents three related challenges. First, native 3DGS is primarily a visual representation, typically static and monolithic, which fails to provide distinct object identities required for planning and manipulation, thereby hindering closed-loop scene state updates~\citep{hu2026worldact,gaussian_grouping}. Second, converting natural language into executable actions requires precise spatial grounding and long-horizon planning. Instructions often leave objects identities and execution orders unspecified, requiring task decomposition based on spatial relationships and action preconditions. Third, expert models differ in their capabilities and motion boundaries. A shared textual goal alone is insufficient for them to disambiguate their respective roles and produce a continuous sequence.

\looseness=-1 We introduce World2HOSI, a stage-based orchestration harness for long-horizon HOSI data synthesis, governed by a closed loop between a structured scene representation and task-specific motion experts. Our framework comprises four foundational modules. First, we propose the \emph{Interactive Scene Package} (ISP), a persistent spatial representation that encapsulates object assets and dynamically tracks evolving spatial relationships. Second, we formulate a scheduler that uses a large language model (LLM) to decompose high-level instructions into structured activity scripts and compile them into numerical representations (MotionIR). Third, to ensure smooth cross-expert transitions, a motion coordinator reconciles motion boundaries and schedules specialized motion experts across different stages. Finally, a physics optimizer applies residual corrections to enforce physical plausibility, writing updated object transformations back to the ISP to condition subsequent interactions. Extensive experiments show that World2HOSI achieves the highest task completion and interaction quality on the evaluated tasks. Further validated by human preference in our user study, World2HOSI provides an effective engine for long-horizon motion planning and generation in complex environments.

\Needspace{4\baselineskip}
In summary, our contributions are threefold: 
\begin{itemize}[leftmargin=1.2em,labelsep=.5em,topsep=3pt,itemsep=2pt,parsep=0pt,partopsep=0pt]
% We present the first multi-expert orchestration harness for long-horizon HOSI data synthesis. This framework resolves stages against this state and compiles expert-specific conditions, enabling complementary capabilities to be used without training an additional
% unified motion generator.
\item \looseness=-1 We present the first multi-expert orchestration harness for long-horizon HOSI data synthesis. The harness grounds each task stage in a shared, continuously updated scene state and constructs expert-specific conditioning inputs, enabling the coordinated use of complementary motion experts without training an additional unified motion generator.
\item \looseness=-1 We introduce a structured scene-state representation for 4D HOSI generation that associates each object's renderable assets and interaction geometry with a persistent identity, enabling heterogeneous motion experts to operate on a shared, consistently updated world state.
% We introduce a structured scene-state substrate for 4D HOSI generation, linking renderable scenes, object interaction geometry, and persistent object identities so that heterogeneous experts share a consistent account of the world.
\item \looseness=-1 We establish a closed-loop generation mechanism that carries state across motion boundaries and writes action outcomes back to the world, allowing subsequent targets to follow the environmental changes induced by earlier actions.
\end{itemize}

%% file: sections/02_related_work.tex
\input{figures/fig02_pipeline}

\section{Related Work}
\label{sec:related-work}

\noindent \looseness=-1 \textbf{From Renderable Scenes to Interactive Worlds.} 3D Gaussian Splatting supports high-quality reconstruction with an explicit Gaussian representation~\citep{kerbl3Dgaussians}, providing visually realistic 3D environments. Making these visual assets interactive additionally requires object-level editability. WorldAct uses object segmentation and geometry construction to make objects in monolithic 3D worlds independently editable and manipulable~\citep{hu2026worldact}. InteriorGS provides indoor Gaussian scenes with semantic annotations and structural information~\citep{InteriorGS2025}, while SAGE-3D builds on these assets to provide mesh representations~\citep{miao2026towards}. Together, these efforts supply assets for scene understanding and interaction research. EgoSim~\citep{hao2026egosim} maintains an evolving 3D world state for egocentric interaction video generation. InHabit demonstrates the potential of reusing foundation-model priors for automated scene-aware human data synthesis, although its outputs are primarily static poses~\citep{kister2026inhabit}. This motivates extending such generative priors to dynamic 4D interaction synthesis.

\noindent \looseness=-1 \textbf{Human-Object and Human-Scene Interaction Generation.} HSI and HOI methods establish distinct motion priors. HSI methods use static environments as geometric constraints, progressively combining obstacle-avoiding navigation with furniture interactions such as sitting and lying down. DIMOS~\citep{zhao2023synthesizing} is a representative example. Complementing this line, MDM~\citep{tevet2023human} uses diffusion for text-conditioned human motion generation. OmniControl~\citep{xie2024omnicontrol} and Kimodo~\citep{Kimodo2026} further support control signals including text, full-body keyframes, and 2D paths, providing priors for diverse motion along specified routes.

\looseness=-1 HOI methods focus on coordination between humans and movable objects. OMOMO~\citep{li2023object} synthesizes full-body manipulation from object trajectories, using intermediate hand positions to strengthen hand-object contact constraints. CHOIS~\citep{li2023controllable} jointly generates human and object motion from text, initial states, and sparse object waypoints, using sampling guidance to improve contact consistency. HOI-FHLI~\citep{hoifhli} further integrates navigation and manipulation generation into a single system, providing more comprehensive priors for long-horizon manipulation. PrimHOI~\citep{jia2025primhoi} composes reusable contact primitives for complex HOI.

\looseness=-1 These conditioning interfaces allow World2HOSI to treat existing models as an orchestrated capability set. A high-level scheduler decomposes tasks, invokes models, and coordinates outputs. Compatible experts can introduce new capabilities without requiring one base model to master every locomotion and interaction type.

\noindent \looseness=-1 \textbf{Joint Human-Object-Scene Interaction Generation.} Several methods pursue joint HOSI generation. HOSIG~\citep{yao2026hosig} incorporates scene constraints into navigation and fine-grained manipulation, while InteractMove~\citep{cai-etal-2025-interactmove} addresses interactions in scenes with movable objects. InfBaGel~\citep{zou2026infbagel} updates dynamic scene perception during generation and mitigates data scarcity through hybrid training on HOI and HSI data. These methods highlight environmental changes and data coverage as central issues. Physical imitation and reference-trajectory refinement~\citep{wang2023physhoi,xu2025intermimic,zhao2026real2sim} further provide contact-driven execution for kinematic sequences, allowing them to satisfy physical dynamics. However, these methods all face data scarcity, which also limits the generalization of their learned priors.

%\looseness=-1 World2HOSI addresses joint interaction data scarcity through synthesis without training an additional unified HOSI generator. By multi-expert orchestration, it reduces dependence on complete HOSI training examples and provides a data-construction pathway for future joint interaction models.

%% file: figures/fig02_pipeline.tex
\begin{figure*}[tb]
\centering
\includegraphics[width=\textwidth]{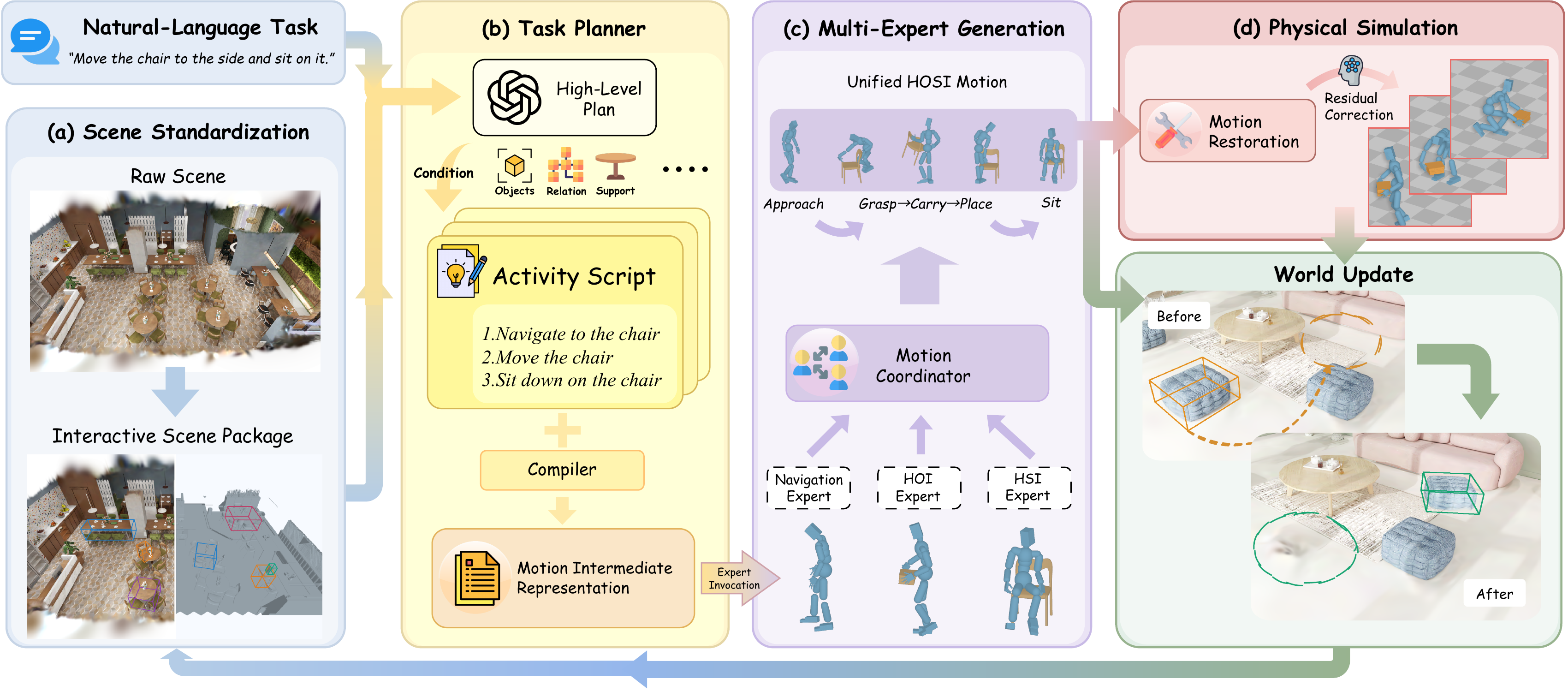}
\caption{World2HOSI closes the loop around a persistent scene state. (a) The ISP organizes background and object assets. (b) Task Planner grounds each action stage in the current state. (c) Experts generate motion, whose outcomes update the ISP before the next invocation. (d) VLM-guided motion restoration selectively refines implausible hand poses during object interactions, followed by physics-based refinement. State write-back keeps subsequent targets consistent with earlier actions. %
}
\label{fig:pipeline}
\end{figure*}

%From a reconstructed static 3DGS indoor scene and natural language instruction, World2HOSI generates long-horizon HOSI through task orchestration, expert invocation, and world updates. (a) Scene standardization builds an Interactive Scene Package (ISP), maintaining the background, object assets, and states as a shared environment. (b) Task planning produces an Activity Script specifying action order, objects, and target regions. Before each stage, the compiler combines the script with current scene and human states to produce MotionIR conditions: paths, poses, and interaction targets. (c) Multi-expert generation invokes navigation, HOI, and HSI experts by stage. A connector adapts segment boundaries, composing complementary capabilities into continuous motion. (d) Physics simulation can refine this reference through bounded human-motion residuals and learn contact-driven manipulation. Outcomes are written back to the ISP, updating object poses, support regions, and obstacles for subsequent conditioning. This loop couples motion orchestration with world updates, allowing later experts to inherit earlier motion and environmental changes.

%% file: sections/03_method.tex
\section{Method}
\label{sec:method}

\looseness=-1 World2HOSI is a multi-model orchestration framework that generates human-object-scene interactions from 3D Gaussian scenes and text instructions. It integrates four core modules, as illustrated in Figure~\ref{fig:pipeline}: Interactive Scene Package (ISP) for persistent spatial scene state, an LLM-based scheduler for script decomposition and compilation, a motion coordinator for cross-expert motion coordination, and a physics optimizer for residual refinement.

\subsection{Interactive World Representation}
\label{sec:world}

\looseness=-1 \textbf{Representation.} While native 3DGS provides a photorealistic collection of Gaussians~\citep{kerbl3Dgaussians}, it lacks the object-level abstraction required for task planning, physical interaction, and persistent state. 
Therefore, we introduce the \emph{Interactive Scene Package} (ISP), a unified structure with persistent object identities. Specifically, each object instance encapsulates visual Gaussian assets, local collision geometry, semantic attributes, a world transformation, and physical properties, while the scene maintains a static background, support regions, object relationships, and navigable space. Rendering and interaction use separate representations aligned by shared object transformations. Gaussian assets provide visual appearance, while geometric proxies support collision checking and path planning. We organize these components into a unified scene state \(W\), which serves as the shared world representation for task planning and motion generation:%
\begin{equation}
W = \left\{V, B, (\mathrm{id}_i,G_i,T_i,A_i,S_i)_{i=1}^{N}, M, R\right\}.
\label{eq:world}
\end{equation}
\looseness=-1 Here, $V$ denotes visual assets, $B$ scene geometry, $\mathrm{id}_i$ a unique identifier for each object, $N$ the number of object instances, $G_i$ local object geometry, and $T_i\in\mathrm{SE}(3)$ world transformation. $A_i$ contains manipulability and physical properties such as mass, $S_i$ denotes object-local support regions, $M$ navigation occupancy, and $R$ object and spatial relationships. Movable objects are separated from the static background, with $T_i$ applied consistently to their visual, collision, and semantic geometry. 

\noindent \looseness=-1 \textbf{Scene Update.} Following object pose changes, the ISP updates support regions and navigation occupancy to condition later tasks. Support regions use object-local coordinates, moving or reorienting a chair recomputes its seat's world coordinates through $T_i$, enabling subsequent sitting without manually reannotating the support surface. Navigation occupancy separates static-background and object-instance layers. Old object occupancy is removed and geometry is reinserted at the new pose, so path planning uses the updated obstacles.

\looseness=-1 The ISP underlies subsequent generation and exposes interpretable task conditions, object movability, the supporting object associated with a target region, and whether a location can be approached are all queryable before motion generation. Details on constructing the ISP are provided in Appendix~\ref{app:isp}.

\subsection{Task Compiler and Scene Grounding}
\label{sec:compiler}

\looseness=-1 Natural language typically describes task intent without exact action orders or spatial coordinates. To let different experts interpret tasks in a shared scene, we separate semantic and numerical representations. At the semantic level, instructions are decomposed into ordered actions in a structured \emph{Activity Script}, recording each stage's action type, object identity, target region, and support relationships. Importantly, object and their local support regions remain persistently bound to specific ISP instances across stages, ensuring that subsequent interactions dynamically track updated object poses rather than outdated initial coordinates. Scripts can be generated by an LLM using predefined prompt templates and ISP information, or specified manually.

\looseness=-1 Before each stage $k$, the compiler reads the symbolic stage description in \emph{Activity Script}, turning it into a numerical representation for motion generation. The stage description is written as $a_k=(c_k,o_k,g_k,\rho_k)$, including action type $c_k$, object identity $o_k$, target area $g_k$, and support relations $\rho_k$. Given the current ISP state $W_k$ and the initial human state $h_k$, the compiler produces a \emph{Motion Intermediate Representation} (MotionIR) instance:
\begin{equation}
\begin{aligned}
\mathcal{C}_k&=\operatorname{Compile}(a_k,W_k,h_k)\\
&=\bigl(c_k,o_k,h_k,T_{o_k,k},\mathcal{P}^{h}_k,\mathcal{P}^{o}_k,\mathcal{G}_k,\eta_k\bigr).
\end{aligned}
\label{eq:motionir}
\end{equation}
\looseness=-1 Here, $h_k$ contains the starting root translation and joint rotations, $T_{o_k,k}$ is the selected object's pose, $\mathcal{P}^{h}_k$ and $\mathcal{P}^{o}_k$ denote human path and object waypoints, and $\mathcal{G}_k$ contains world-space target poses or support regions resolved from $g_k$. The metadata $\eta_k$ records coordinate frame, units, temporal sampling, and human representation. The resulting MotionIR instance $\mathcal{C}_k$ is passed to the motion expert, providing an explicit interface between stage-level descriptions and motion generation.

\looseness=-1 Compilation thus grounds textual intent in stage-wise numerical conditions. The two intermediate representations unify task semantics and numerical specifications, keeping scene and task understanding from being delegated entirely to motion experts.

\subsection{Multi-Expert Motion Synthesis}
\label{sec:experts}

\looseness=-1 \textbf{Generation.} Once the current stage has been grounded, the motion coordinator uses its action type to orchestrate experts from a configured mapping. Experts are defined by capability, independently of specific models. Most HOSI tasks can be decomposed into navigation, dynamic HOI (e.g., carrying, pushing, or pulling), and static HSI (e.g., sitting or lying down). The navigation expert generates scene-constrained locomotion, the HOI expert jointly generates human and object motion, and the HSI expert generates interactions with static scene geometry. All receive conditions through the same upstream interface, decoupling task representation from individual generators. The harness is therefore independent of any model or architecture, experts can be replaced and new capabilities added without redesigning the pipeline.

\noindent \looseness=-1 \textbf{Cross-Expert Reconciliation.} While \emph{MotionIR} aligns task semantics and spatial targets, independently generated segments may still exhibit boundary jumps. We therefore introduce a boundary projection mechanism that anchors only the incoming segment's prefix to the preceding terminal state without altering completed motion. Let $(p^{-}, R_j^{-})$ denote the preceding final root position and rotation of joint $j$, and let $(p^{+}(t), R_j^{+}(t))$ denote the incoming expert's original trajectories at elapsed time $t \geq 0$. Given the initial boundary offsets $\Delta p = p^{-} - p^{+}(0)$ and $\Delta R_j = R_j^{-}\left(R_j^{+}(0)\right)^{-1}$, we smoothly relax them over a window $\tau$ via a quintic smoothstep weight $s(u)\in[0,1]$ ($u=\min(t/\tau,1)$) to obtain the corrected trajectories $\widetilde{p}^{+}(t)$ and $\widetilde{R}_{j}^{+}(t)$:
\begin{align}
\widetilde{p}^{+}(t) &= p^{+}(t)+\left(1-s(u)\right)\Delta p,
\label{eq:translation-transition}\\
\widetilde{R}_{j}^{+}(t) &= \operatorname{Slerp}\!\left(\Delta R_j,I,s(u)\right)R_j^{+}(t),
\label{eq:rotation-transition}
\end{align}
\looseness=-1 where $\operatorname{Slerp}$ denotes spherical linear interpolation of rotations and $I$ is the identity rotation. This mechanism enforces positional and rotational continuity at $t=0$ while restricting trajectory deviations to the initial window $\tau$, fully recovering the incoming expert's original motion prior thereafter. Such projection mechanism guarantees temporal continuity with the existing motion state.

\subsection{Physics-Based Motion Refinement}
\label{sec:physics}

\looseness=-1 Kinematic motion priors do not explicitly enforce gravity, friction, or sustained contact, so even the motion expert's output may contain physical violations. While physics-based tracking can resolve minor penetrations and floating, it often fails outright under severely misaligned hand-object grasps, as local residual adjustments cannot bridge large inappropriate contact. We therefore introduce a two-stage physical refinement pipeline comprising VLM-guided motion restoration followed by RL-based residual refinement. First, we sample multiple close-up hand viewpoints and leverage the visual-semantic priors of a Vision-Language Model (VLM) to reposition finger joints onto plausible grasping regions, while keeping the remaining body joints frozen to preserve global motion intent. Using this restored kinematic reference, an RL controller~\citep{zhao2026real2sim,zhao2025resmimic} then makes bounded residual adjustments to joint targets executed by proportional-derivative (PD) controllers. By balancing reference tracking with motion smoothness, the controller yields a physically executable realization while preserving interaction intent. Detailed control laws, reward terms, and optimization settings are provided in Appendix~\ref{app:physics-details}.

%% file: sections/04_experiments.tex
\section{Experiments}
\label{sec:experiments}

\looseness=-1 We extensively evaluate World2HOSI through quantitative comparisons, perceptual user studies, and component ablations. The evaluation covers generation quality, stitching effectiveness, and physical plausibility.

\input{figures/fig03_shared_task_comparison}
\input{layouts/shared_tasks_and_user_study}

\subsection{Comparison on Long-Horizon HOSI Tasks}
\label{sec:shared-tasks}

\looseness=-1 Our harness targets long-horizon HOSI. We assess the generated data through quantitative measures of task completion and geometric quality, together with human perceptual evaluation.

\noindent \looseness=-1\textbf{Datasets.} To evaluate the effectiveness of our method on HOSI tasks, we construct a benchmark on 10 Interactive Scene Package (ISP) scenes. These scenes span diverse indoor layouts populated with rich interactable objects. Within these environments, we design 20 long-horizon, human-level HOSI task instructions systematically categorized into five functional groups: (1) dynamic object manipulation, (2) support-based scene interactions, (3) combined manipulation and support interactions, (4) state-dependent tasks, and (5) continuous multi-object manipulation. Each task is evaluated across three random seeds, generating 60 motion sequences in total. All compared methods are tested under identical initial scene configurations, object geometries, and target conditions. More details on scene construction and complete task specifications are provided in Appendices~\ref{app:isp} and~\ref{app:tasks}.

\noindent \looseness=-1 \textbf{Metrics.} We evaluate motion quality and interaction from three perspectives: task completion, interaction plausibility, and human motion quality. For task completion, waypoint following errors $T_{xy}$~\citep{hoifhli,li2023controllable} measures how the human root trajectory deviates from the planned route during approach, carrying after pickup, and subsequent approaches. The task progress score $S_t$ decomposes instructions into ordered stages and continuously scores partial completion. Carrying, pushing, and pulling check final human and object distances to their targets; sitting and lying check target pose changes. Later stages receive credit only after prerequisite actions are observed. The geometry score $S_g$ aggregates geometric quality metrics. All metrics are calculated before physics simulation, unless otherwise specified. The detailed scoring rules are provided in Appendix~\ref{app:metrics}. 

\looseness=-1 For interaction plausibility, we measure the contact ratio during object motion (OCP, \%)~\citep{zou2026infbagel,li2023controllable}, severe-penetration frame ratio ($P_f$, \%)~\citep{zhao2023synthesizing,zou2026infbagel}, using a penetration-depth threshold of 2 cm. Non-contact object displacement (NCOD, cm), measured as the maximum object-surface displacement before first hand contact, and mean object-scene penetration ($P_{\mathrm{os}}$)~\citep{zhao2023synthesizing,zou2026infbagel}, assessing how plausibly a human could execute the interaction. 

\looseness=-1 For human motion quality, we measure mean human-object penetration ($P_{\mathrm{ho}}$)~\citep{zou2026infbagel}, foot sliding score (FS)~\citep{hoifhli,zou2026infbagel}, and foot-ground deviation ($O_{\mathrm{feet}}$).

\paragraph{Baselines.}
\looseness=-1 Although recent years have witnessed significant advances in motion generation, the complex nature of HOSI poses unique challenges, leaving it supported by only two existing baselines, i.e. InfBaGel~\citep{zou2026infbagel} and HOSIG~\citep{yao2026hosig}. Meanwhile, to validate the benefit of our multi-expert orchestration, we also compare against a single-expert variant using HOI-FHLI. For fairness, our harness compiles shared target objects, target regions, approach positions, and navigation paths, ensuring identical task semantics, scenes, and object geometry across methods. Navigation paths are constructed using A* on clearance-inflated occupancy grids, followed by collision-checked simplification and smoothing. 

\looseness=-1 Table~\ref{tab:shared-tasks} shows the highest task progress and aggregate geometry scores for World2HOSI. Our method follows the planned tasks, completing intermediate operations more fully while approaching targets. InfBaGel reduces foot sliding but has lower progress and contact, indicating that higher-quality human motion alone is insufficient for long-horizon HOSI. HOSIG achieves relatively high completion but 85.44\% severe-penetration frames proximity or contact alone does not ensure plausibility, exposing limitations in joint human-object generation. Our balance of completion and geometry supports the benefits of unified closed-loop state update mechanism. %

\looseness=-1 This advantage is further corroborated by the qualitative comparisons in Figure~\ref{fig:qualitative}. In these depicted tasks, InfBaGel fails to maintain valid contact or complete stage transitions, whereas HOSIG incurs severe human-object and scene penetration. By contrast, World2HOSI generates plausible interactions that accurately adapt to dynamic scene state updates.

\input{tables/tab02_infbagel_benchmark}
\input{figures/fig05_infbagel_comparison}

\subsection{Experiment on InfBaGel benchmark}
\label{app:benchmark}

\looseness=-1 To evaluate cross-scene generalization and directly compare against existing methods, 
we evaluate the performance of our method on the InfBaGel benchmark~\citep{zou2026infbagel}, which contains 469 tasks with seven object categories in 67 indoor scenes.

\noindent \looseness=-1 \textbf{Metrics.} Following the evaluation protocol of InfBaGel, we assess the generated motions from three perspectives. Regarding task completion, we measure the final distances of the human ($T_h$) and the object ($T_o$) from their respective targets, as well as the task success rate ($S\%$), a task is considered successful only if both distances are below 10 cm. For motion and interaction quality, we evaluate foot sliding (FS), contact percentage ($C\%$), and human-body penetration ($P_{body}$). Regarding scene awareness, we measure scene penetration for both the human and the object, calculating the average and maximum penetration vertex depths ($P_{mean}$, $P_{max}$), as well as the percentage of frames in which penetration occurs ($P_f\%$).

\noindent \looseness=-1 \textbf{Quantitative Results.} As reported in Table~\ref{tab:infbagel-benchmark}, World2HOSI achieves the best results across most metrics on the InfBaGel benchmark. By substantially improving hand and object contact $C\%$ while suppressing body penetration $P_{\text{body}}$, scene penetration, and terminal placement errors, our method boosts the overall task success rate $S\%$ to $85.93\%$. The results validate our framework's effectiveness in executing complex long-horizon tasks with strong physical plausibility.

\noindent \looseness=-1 \textbf{Qualitative Comparison.} Figure~\ref{fig:infbagel} presents a visual comparison between InfBaGel and World2HOSI across two representative manipulation tasks on the InfBaGel benchmark. As highlighted by the red circles, InfBaGel frequently suffers from severe physical artifacts during the HOI process. Specifically, InfBaGel exhibits pronounced hand-object interpenetration and misaligned grasping poses, struggles to maintain stable contact during object lifting and placement. In contrast, World2HOSI consistently generates natural, plausible grasping postures along with smooth full-body motion trajectories. These qualitative results confirm that our multi-expert harness effectively maintains plausibility throughout HOSI interactions.

\input{figures/fig06_physics_refinement}
\input{figures/fig07_long_horizon}

\subsection{Ablation Study}
\label{sec:ablation}

\paragraph{Cross-expert Reconciliation.}
\looseness=-1 We evaluate the cross-expert boundary reconciliation by comparing direct concatenation (A0), bilateral translation blending with spherical linear interpolation (A1), and our boundary projection mechanism (A2), which keeps the preceding segment fixed while decaying the initial boundary offsets of the incoming segment over a local window. Within $0.5\,\mathrm{s}$ of each boundary, A2 substantially reduces joint displacement, angular change, and velocity jumps compared to A1, effectively suppressing cross-expert discontinuities without distorting preceding motion segments. Full quantitative results are provided in Table~\ref{tab:reconciliation}.

\input{tables/tab03_physics_ablation}

\Needspace{2\baselineskip}
\paragraph{Physics-Based Residual Refinement.}%
\looseness=-1 We evaluate Physics-Based Residual Refinement on 20 clips selected for their relatively complex contact dynamics, averaging approximately 10\,s and involving objects of diverse shapes and types.

\looseness=-1 Table~\ref{tab:physics-ablation} reports six metrics covering interpenetration and contact: mean human-object penetration ($P_{\mathrm{ho}}$), severe-penetration frame ratio ($P_f$), foot sliding score ($\text{FS}$), foot-ground deviation ($O_{\mathrm{feet}}$), hand–object contact distance ($\text{HOCD}$, cm) and foot skating ratio ($\text{FSR}$), which measures the proportion of frames exhibiting noticeable foot slippage during ground contact~\citep{zeng2025chainhoi}. We observe that, even without physics-based optimization, VLM-guided grasp restoration reduces both human-object penetration and the proportion of frames with severe penetration. Physics-based optimization further enhances physical plausibility by mitigating penetration, foot sliding, and foot-ground deviation. Notably, incorporating VLM-guided grasp restoration into the physics-based optimization also improves foot-ground contact stability. This suggests that inaccurate grasp poses can introduce unstable contacts during physical execution, thereby inducing body imbalance. By rectifying these grasp poses, VLM-guided restoration helps sustain more stable human-object interactions throughout the motion sequence. 

\looseness=-1 Physics-based optimization further improves physical plausibility by reducing reference artifacts, including foot sliding, floating, and interpenetration, as illustrated in Figure~\ref{fig:physics}. 
Contact-driven execution allows local adjustments to the reference motion, helping establish more consistent human-object and foot-ground interactions.

\input{tables/tab04_gpt_comparison}

\subsection{Comparison with GPT-6-Astra}
\label{app:gpt}
\looseness=-1 As frontier models become increasingly capable, we investigate whether a general-purpose model can directly generate long-horizon human–object–scene interactions. We evaluate GPT-6-Astra~\citep{openai2026gpt6astra} as a standalone motion-scripting baseline, conditioned on task instructions, scene geometry, and initial human and object states. GPT-6-Astra writes executable code to jointly synthesize human poses and object trajectories. 

\noindent \looseness=-1 \textbf{Quantitative Results.} To evaluate the effectiveness of our harness, we employ GPT-6-Astra as a standalone baseline to directly generate human–object–scene interaction on the tasks corresponding to our physics-refined sequences. As reported in Table~\ref{tab:gpt}, without structured harness guidance and physical refinement, GPT-6-Astra struggles with accurate task execution and physical realism, which exhibits an excessively high $\text{NCOD}$ alongside notably worse penetration and foot-sliding scores. This underscores the challenge of one-shot joint human--object synthesis, which frequently produces severe contact failures such as mid-air grasping. Conversely, World2HOSI delivers marked improvements in task completion, physical contact consistency, and overall motion quality. Moreover, programmatic motions generated by GPT appear rigid and mechanical, whereas our approach produces substantially more naturalistic kinematics. We include qualitative comparisons in Figure~\ref{fig:GPT} for a more intuitive visual assessment.

\subsection{Long-Horizon Stateful Interaction}
\label{sec:long-horizon}

\looseness=-1 To evaluate how our closed-loop state-update mechanism supports complex, long-horizon HOSI, we test multi-stage tasks with cross-stage state dependencies. As shown in Figure~\ref{fig:long-sequences}, after relocating a chair, drinking coffee, and placing the cup on a table , the human grounds the final sitting motion on the chair's updated pose and support region. Likewise, moving books and repositioning a stool updates the object pose and spatial layout for the subsequent sitting interaction. Both tasks involve subsequent interactions grounded in updated object poses, and our method successfully completes all stages in both cases.

\subsection{User Study}
\label{sec:user-study}

\looseness=-1 Since current quantitative metrics often fail to fully capture visual realism and physical plausibility, we conduct a user study to evaluate the true perceptual quality and naturalness of the generated motions from a human perspective. Using each method on the shared tasks, we generate a total of 24 sequences and organize them into eight comparison videos, each containing our method and two baselines with identical rendering and camera settings. An online questionnaire collects responses from 30 participants, who select the method with the highest task completion and best interaction quality for each comparison. Our method substantially leads the baselines in preference rates on both criteria (Figure~\ref{fig:user-study}).

%% file: figures/fig03_shared_task_comparison.tex
\begin{figure}[tb]
\centering
\includegraphics[width=\linewidth]{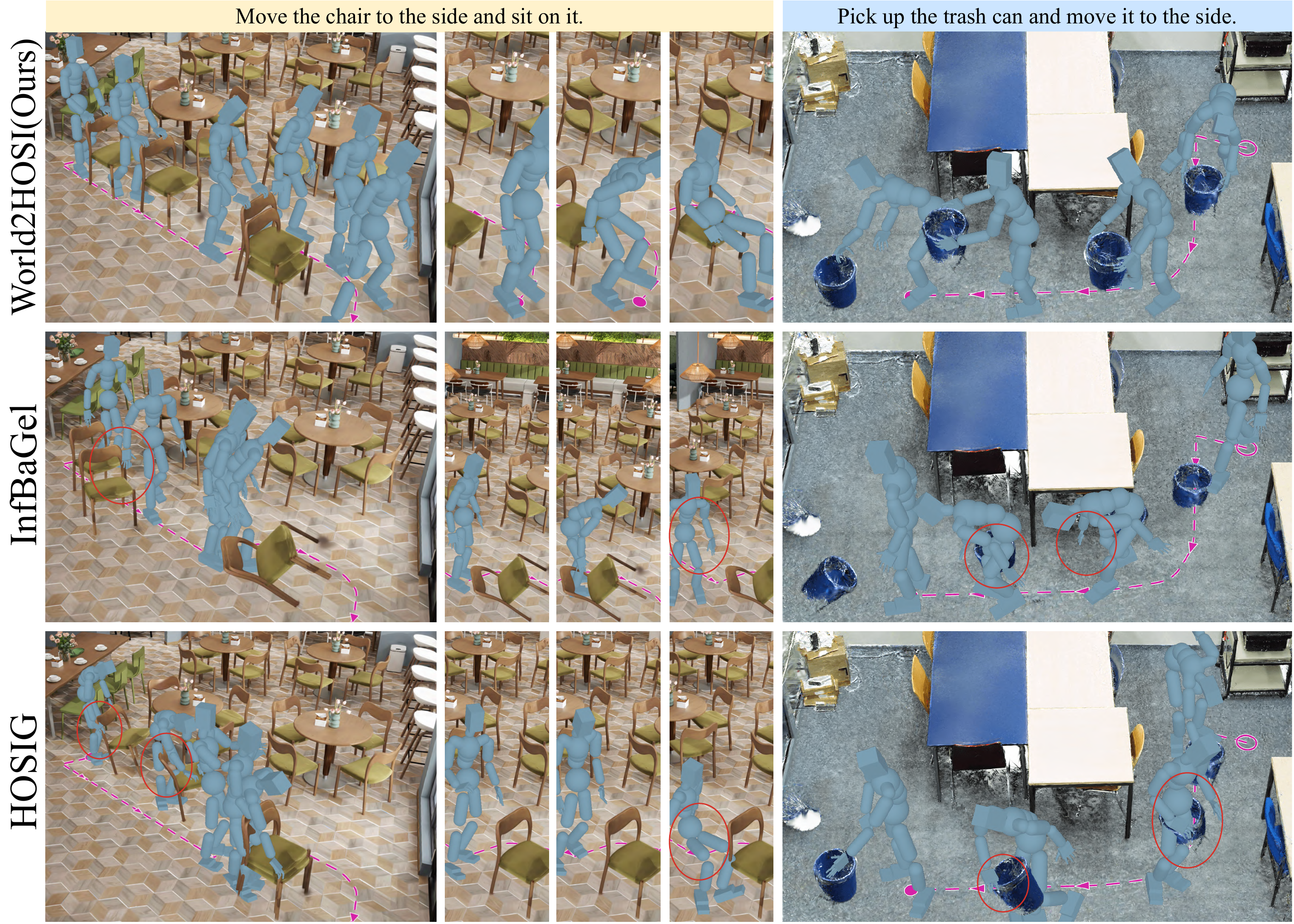}
\caption{Qualitative comparison with external methods. We compare both an object-carrying task and a more complex multi-stage task. }
\label{fig:qualitative}
\end{figure}

%% file: layouts/shared_tasks_and_user_study.tex
% Table 1 and Figure 3 share one row without scaling either table text or chart text.
\begin{table*}[tb]
\centering
\begin{minipage}[t]{356pt}
\vspace{0pt}
\input{tables/tab01_shared_tasks}
\end{minipage}\hfill
\begin{minipage}[t]{130pt}
\vspace{0pt}
\input{figures/fig04_user_study}
\end{minipage}
\end{table*}

%% file: tables/tab01_shared_tasks.tex
\centering
\captionof{table}{Quantitative results on the shared HOSI tasks. Compared with the external baselines, our method achieves superior task completion and interaction quality.}
\label{tab:shared-tasks}

\begingroup
\small
\setlength{\tabcolsep}{3.2pt}
\renewcommand{\arraystretch}{1.0}
\begin{tabular}{lcccccccccc}
\toprule
\multirow{2}{*}{Method} & \multicolumn{3}{c}{Task Completion} & \multicolumn{4}{c}{Interaction Plausibility} & \multicolumn{3}{c}{Human Motion Quality} \\
\cmidrule(lr){2-4}\cmidrule(lr){5-8}\cmidrule(lr){9-11}
& $T_{xy}\downarrow$ & $S_t\uparrow$ & $S_g\uparrow$ & $\mathrm{OCP}\uparrow$ & $P_f\downarrow$ & $\mathrm{NCOD}\downarrow$ & $P_{\mathrm{os}}\downarrow$ & $P_{\mathrm{ho}}\downarrow$ & $\mathrm{FS}\downarrow$ & $O_{\mathrm{feet}}\downarrow$ \\
\midrule
InfBaGel & 36.02 & 42.54 & 44.61 & 8.52 & 47.99 & 146.91 & 2.201 & 2.433 & \textbf{0.048} & 4.77 \\
HOSIG & 44.53 & 65.33 & 48.16 & 78.14 & 85.44 & 2.97 & 2.077 & 2.069 & 0.823 & 5.27 \\
\midrule
HOI-FHLI & 23.14 & 62.54 & 64.68 & 77.29 & 43.36 & 2.88 & 1.105 & 1.554 & 0.467 & 4.61 \\
Ours & \textbf{17.88} & \textbf{76.93} & \textbf{75.61} & \textbf{84.94} & \textbf{24.92} & \textbf{2.08} & \textbf{0.708} & \textbf{1.309} & 0.255 & \textbf{4.27} \\
\bottomrule
\end{tabular}%
\endgroup

%% file: figures/fig04_user_study.tex
% Included by layouts/shared_tasks_and_user_study.tex.
\centering
\includegraphics[width=\linewidth]{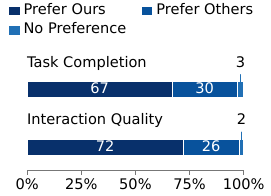}
\captionof{figure}{Comparison with the baselines in the perceptual user study.}
\label{fig:user-study}

%% file: tables/tab02_infbagel_benchmark.tex
\begin{table*}[tb]
\centering
\caption{Comparison on the InfBaGel benchmark. Our method achieved the best results across most metrics.}
\label{tab:infbagel-benchmark}
\begingroup
\small
\setlength{\tabcolsep}{4pt}
\begin{tabular*}{\linewidth}{@{\extracolsep{\fill}}lcccccccccccc@{}}
\toprule
\multirow{2}{*}{Method} & \multicolumn{3}{c}{Task accuracy} & \multirow{2}{*}{FS$\downarrow$} & \multicolumn{2}{c}{HO interaction} & \multicolumn{3}{c}{HS penetration $\downarrow$} & \multicolumn{3}{c}{OS penetration $\downarrow$} \\
\cmidrule(lr){2-4} \cmidrule(lr){6-7} \cmidrule(lr){8-10} \cmidrule(lr){11-13}
 & $T_h\downarrow$ & $T_o\downarrow$ & $S\%\uparrow$ & & $C\%\uparrow$ & $P_{body}\downarrow$ & $P_{mean}$ & $P_{max}$ & $P_f\%$ & $P_{mean}$ & $P_{max}$ & $P_f\%$ \\
\midrule
InfBaGel   & 4.37  & 7.94 & 81.45 & \textbf{0.15} & 76.96 & 5.05 & \textbf{3.17} & 37.09 & 26.74 & 12.45 & \textbf{93.37} & 22.32 \\
HOSIG   & 23.54 & 6.00 & 22.39 & 0.17 &  51.92 & 4.33 & 54.79 & 221.53 & 90.22 & 86.34 & 450.19 & 32.26 \\
World2HOSI   & \textbf{4.04} & \textbf{1.62} & \textbf{85.93} & 0.45 & \textbf{88.48} & \textbf{1.96} & 3.75 & \textbf{34.59} & \textbf{23.20} &  \textbf{7.53} & 94.54 & \textbf{21.13} \\
\bottomrule
\end{tabular*}
\endgroup
\end{table*}

%% file: figures/fig05_infbagel_comparison.tex
\begin{figure*}[tb]
\centering
\includegraphics[width=\linewidth]{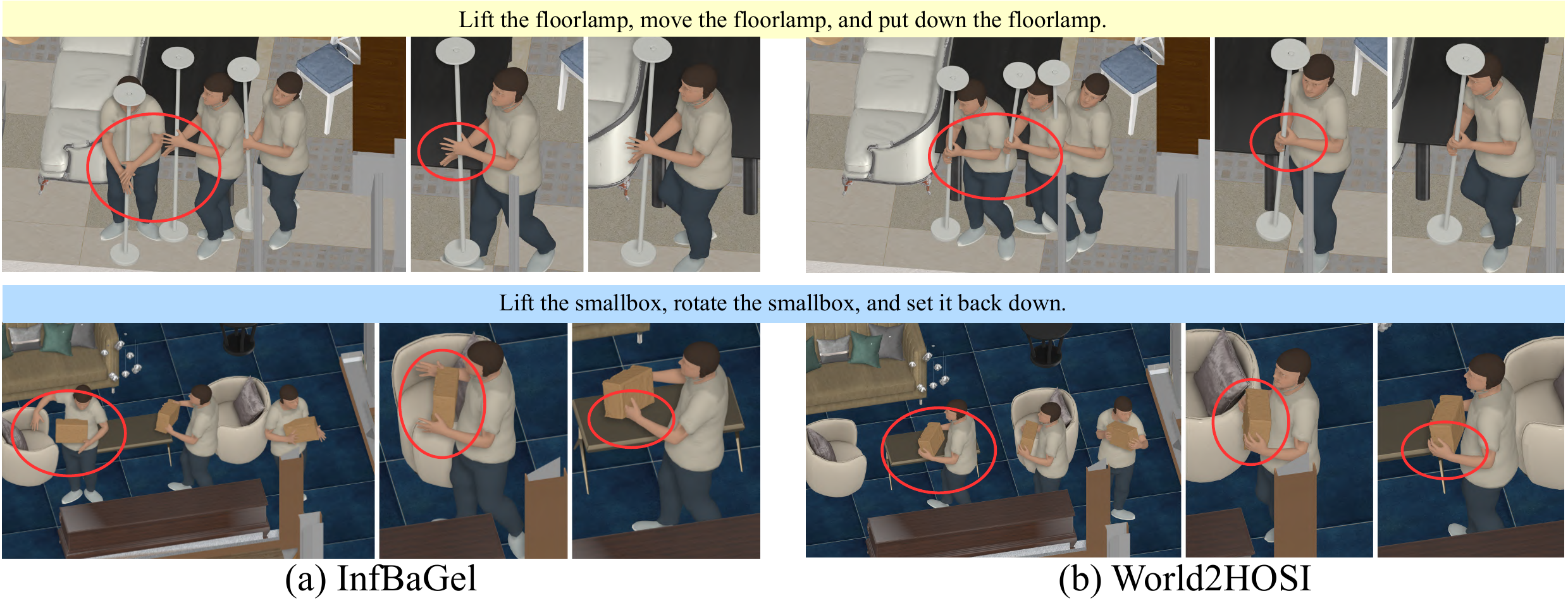}
\caption{Qualitative comparison between (a) InfBaGel and (b) World2HOSI. Red circles highlight key hand–object contact regions.}
\label{fig:infbagel}
\end{figure*}

%% file: figures/fig06_physics_refinement.tex
\begin{figure}[tb]
\centering
\includegraphics[width=\linewidth]{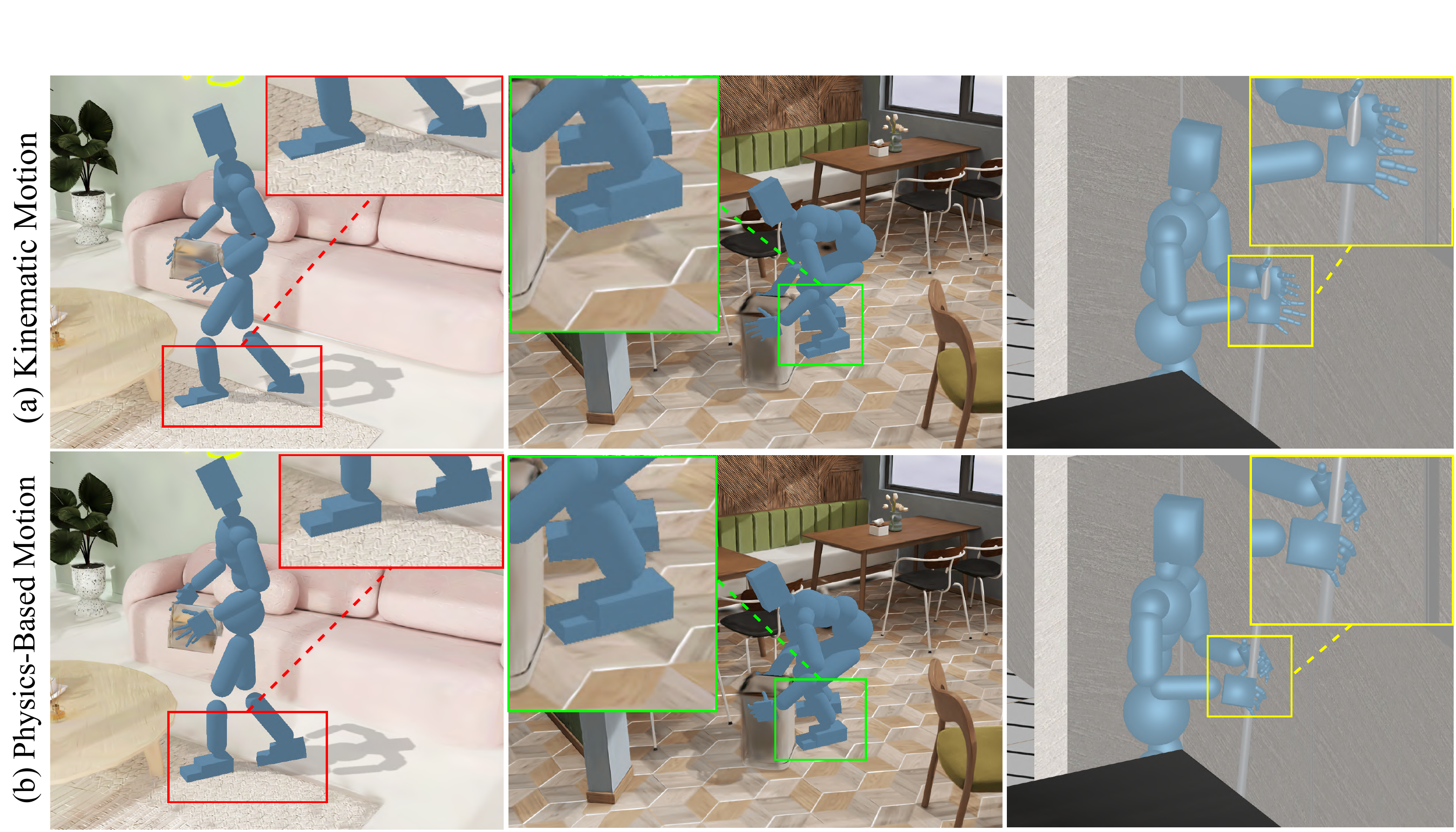}
\caption{Qualitative comparison between kinematic reference motion and physics-based motion. The residual refinement policy effectively removes artifacts such as floating and interpenetration, while also improving grasping poses.}
\label{fig:physics}
\end{figure}

%% file: figures/fig07_long_horizon.tex
\begin{figure*}[tb]
\centering
\includegraphics[width=\linewidth]{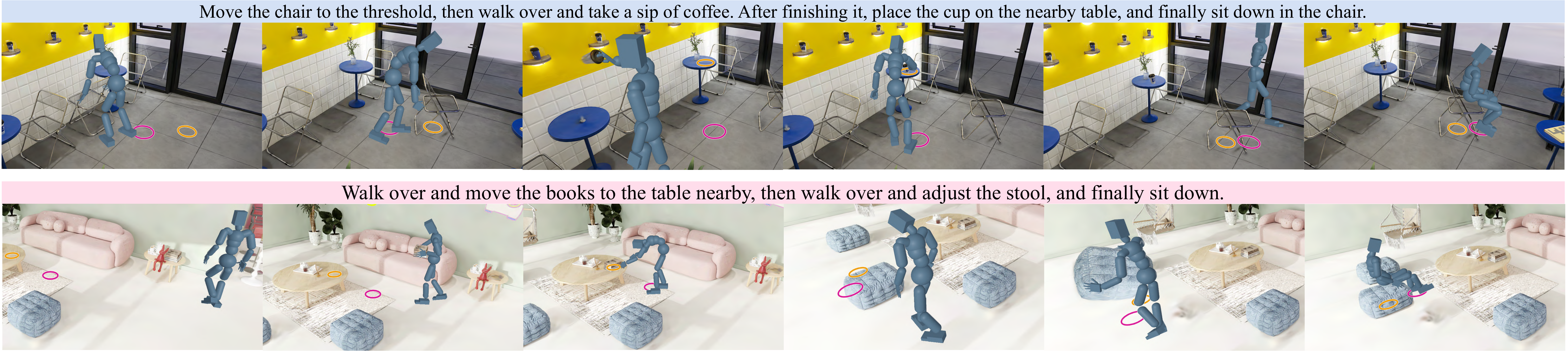}
\caption{Long-horizon interaction with persistent scene state. Top: relocating a chair, drinking coffee, and returning to sit. Bottom: moving books, repositioning a stool, and sitting. Updated object poses and support regions in the ISP guide subsequent interactions.}
\label{fig:long-sequences}
\end{figure*}

%% file: tables/tab03_physics_ablation.tex
\begin{table}[tb]
\centering
\caption{Quantitative ablation on physical refinement and motion quality.}
\label{tab:physics-ablation}

\begingroup
\small
\setlength{\tabcolsep}{1.6pt}
\renewcommand{\arraystretch}{1.0}
\begin{tabular*}{\linewidth}{@{\extracolsep{\fill}}lcccccc@{}}
\toprule
Method & $P_{\text{ho}}\downarrow$ & $P_f\downarrow$ & $\text{FSR}\downarrow$ & $\text{FS}\downarrow$ & $O_{\text{feet}}\downarrow$ & $\text{HOCD}\downarrow$ \\
\midrule
Kinematic & 5.408 & 28.78 & 0.104 & 0.345 & 2.02 & 0.457 \\
Kinematic + VLM & 3.169 & 13.64 & 0.104 & 0.345 & 2.02 & 0.031 \\
Physics w/o VLM & 0.643 & 4.70 & 0.076 & 0.083 & 0.28 & 0.348 \\
Physics-based (Ours) & \textbf{0.242} & \textbf{3.36} & \textbf{0.038} & \textbf{0.052} & \textbf{0.12} & \textbf{0.016} \\
\bottomrule
\end{tabular*}
\endgroup\end{table}

%% file: tables/tab04_gpt_comparison.tex
\begin{table}[tb]
\centering
\caption{Quantitative comparison between World2HOSI and GPT-6-Astra on shared HOSI tasks.}
\label{tab:gpt}
\begingroup
\small
\setlength{\tabcolsep}{1.7pt}
\renewcommand{\arraystretch}{0.92}
\begin{tabular*}{\linewidth}{@{\extracolsep{\fill}}lcccccc@{}}
\toprule
\multirow{2}{*}{Method} & \multicolumn{3}{c}{Task Completion} & \multicolumn{3}{c}{Motion Quality} \\[-1pt]
\cmidrule(lr){2-4}\cmidrule(lr){5-7}
& $T_{xy}\downarrow$ & $S_t\uparrow$ & $S_g\uparrow$ & $P_{\mathrm{ho}}\downarrow$ & $\mathrm{FS}\downarrow$ & $O_{\mathrm{feet}}\downarrow$ \\
\midrule
GPT-6-Astra & 34.15 & 75.79 & 55.71 & 1.614 & 0.162 & 2.94 \\
Ours & \textbf{9.06} & \textbf{76.37} & \textbf{95.74} & \textbf{0.242} & \textbf{0.052} & \textbf{0.12} \\
\midrule
\end{tabular*}\par\nointerlineskip
\begin{tabular*}{\linewidth}{@{\extracolsep{\fill}}lcccc@{}}
\multicolumn{1}{l}{\multirow{2}{*}{}} & \multicolumn{4}{c}{Interaction Plausibility} \\[-1pt]
\cmidrule(lr){2-5}
& $\mathrm{OCP}\uparrow$ & $P_f\downarrow$ & $\mathrm{NCOD}\downarrow$ & $P_{\mathrm{os}}\downarrow$ \\
\midrule
GPT-6-Astra & 44.54 & 51.92 & 39.39 & \textbf{0.417} \\
Ours & \textbf{94.22} & \textbf{3.36} & \textbf{0.15} & 0.590 \\
\bottomrule
\end{tabular*}
\endgroup
\end{table}

%% file: sections/05_conclusion.tex
\section{Conclusion}
\label{sec:conclusion}

\looseness=-1 We presented World2HOSI, an orchestration framework for long-horizon HOSI data generation. It transforms static 3D Gaussian scenes into continuously updated interactive worlds. Through the \emph{Interactive Scene Package}, task compilation, and projection-based stitching, navigation, human-object, and human-scene models generate continuous motion under shared state, with object motion affecting subsequent task conditions. Bounded residual optimization further refines the kinematic reference for physical plausibility. This loop establishes the first systematic multi-expert orchestration framework for complex, long-horizon HOSI tasks, offering a new pathway to alleviate joint interaction data scarcity and a reusable engine for future research and extensions.

%% file: sections/06_ai_usage.tex
We used a large language model (LLM) as an auxiliary tool to polish the manuscript and search for information. Specifically, we employed Gemini 3.8 to assist with translation, language polishing, and the refinement of sentence flow. Additionally, we used GPT-6 to help search for literature related to our work. The authors take full responsibility for the final content, including text generated or refined with the LLM.

%% file: sections/appendix.tex
\section{Appendix}
\label{app:appendix}

\subsection{ISP Construction Details}
\label{app:isp}

An ISP can be constructed either from annotated scene assets or from a monolithic 3DGS reconstruction~\citep{kerbl3Dgaussians}. For annotated scenes from ScanNet++~\citep{yeshwanth2023scannet++} and InteriorGS~\citep{InteriorGS2025}, we use the scanned meshes, instance annotations, and initial 3D bounding boxes to select informative views. Text prompts and projected bounding boxes are then provided to SAM3~\citep{carion2025sam3segmentconcepts} to obtain multi-view object masks. These segmentation results are fused using camera poses, scan depth, and Gaussian projection contributions to extract object-level mesh and Gaussian assets.

For unannotated 3DGS scenes, we adopt a WorldAct-style pipeline~\citep{hu2026worldact}. Qwen3.6-Plus~\citep{qwen36plus} is used to identify separable objects and evaluate candidate views, followed by multi-view segmentation with SAM3. We then determine the instance assignment of 3D Gaussians using camera calibration, occlusion relationships, and Gaussian projection contributions, while unassigned Gaussians are retained as the static background. The required meshes are obtained from existing scans when available, or reconstructed using Gaussian Wrapping~\citep{gomez2026blobsspokeshighfidelitysurface}.

Finally, the decoupled visual assets and collision geometry are organized in a shared metric coordinate frame, with persistent object identities, bounding boxes, and initial occupancies recorded for each instance. Support regions are extracted from mesh surfaces and stored in object-local coordinates. Category-level priors provided by an LLM are further used to initialize basic properties such as manipulability, mass, and friction. Before finalizing the ISP, manual inspection and correction can be applied to segmentation, alignment, functional orientation, and support geometry.

\subsection{Complete Task Definitions}
\label{app:tasks}

This appendix lists the semantic structure of the tasks used in the main experiments. 

\subsubsection{Task Instances and State Dependencies}
\label{app:task-instances}

\newcommand{\taskgroup}[1]{\vspace{4pt}\noindent\textbf{#1}\par\vspace{2pt}}
\newenvironment{tasklist}{%
  \begin{list}{}{%
    \setlength{\labelwidth}{2.2em}%
    \setlength{\leftmargin}{2.6em}%
    \setlength{\itemsep}{0pt}%
    \setlength{\parsep}{0pt}%
    \setlength{\topsep}{0pt}%
  }%
}{\end{list}}

\taskgroup{Dynamic object manipulation (T01-T06).}
\begin{tasklist}
  \item[T01] Approach a box $\rightarrow$ carry it $\rightarrow$ place it.
  \item[T02] Approach a stack of books $\rightarrow$ carry it $\rightarrow$ place it on a table.
  \item[T03] Approach another box along a specified route $\rightarrow$ carry it $\rightarrow$ place it.
  \item[T04] Approach a book $\rightarrow$ carry it $\rightarrow$ place it.
  \item[T05] Approach the target object $\rightarrow$ pull it to a specified region.
  \item[T06] Approach a chair $\rightarrow$ push it once to a specified region.
\end{tasklist}

\taskgroup{Support-based scene interactions (T07-T10).}
\begin{tasklist}
  \item[T07] Navigate $\rightarrow$ sit down.
  \item[T08] Navigate along a specified route $\rightarrow$ sit down.
  \item[T09] Navigate to a lying surface $\rightarrow$ lie down.
  \item[T10] Navigate to another seat $\rightarrow$ sit down.
\end{tasklist}

\taskgroup{Combined manipulation and support interactions (T11-T14).}
\begin{tasklist}
  \item[T11] Carry a chair $\rightarrow$ sit on it at its new location.
  \item[T12] Carry a supporting object $\rightarrow$ sit on it at its new location.
  \item[T13] Carry a stack of books $\rightarrow$ navigate to a seat $\rightarrow$ sit down.
  \item[T14] Carry a book $\rightarrow$ navigate to another seat $\rightarrow$ sit down.
\end{tasklist}

\taskgroup{State-dependent tasks (T15-T18).}
\begin{tasklist}
  \item[T15] Move the same box from A $\rightarrow$ B $\rightarrow$ C through two consecutive carrying actions.
  \item[T16] Move the same stack of books from A $\rightarrow$ B $\rightarrow$ A, returning it after carrying.
  \item[T17] Pull the target object out of the way $\rightarrow$ navigate through the updated region.
  \item[T18] Relocate the target object $\rightarrow$ navigate through the updated region.
\end{tasklist}

\taskgroup{Continuous manipulation on multiple objects (T19-T20).}
\begin{tasklist}
  \item[T19] Carry a stack of books $\rightarrow$ carry another object.
  \item[T20] Carry the first chair $\rightarrow$ carry the second chair.
\end{tasklist}

\subsubsection{Shared Conditions and Model Invocation}
\label{app:shared-conditions}

Our shared-task implementation uses Kimodo~\citep{Kimodo2026} for navigation, HOI-FHLI~\citep{hoifhli} for dynamic object interactions, and DIMOS~\citep{zhao2023synthesizing} for static interaction. For the diversity extension, we additionally use LINGO's scene interaction capability~\citep{jiang2024autonomous}. All outputs are converted to a shared world coordinate frame, metric units, and a common human representation. Each task-seed pair is associated with identical scene and object geometry, initial states, and target conditions across methods.

InfBaGel and HOSIG use their respective pretrained generators~\citep{zou2026infbagel,yao2026hosig}. Their adaptation layers only perform geometry and coordinate conversion, condition encoding, and output conversion back to the shared representation. To accommodate the baselines' limitations on long-horizon tasks, the same baseline may be invoked stage by stage for complex tasks, with each stage inheriting the actual generated state of the previous stage and the resulting segments concatenated.

\input{figures/figS1_infbagel}

\subsection{Full Quantitative Results on Cross-expert Reconciliation Ablation.}
\label{sec:crossexpert}

\input{tables/tabS1_reconciliation}

Table~\ref{tab:reconciliation} provides the full quantitative results for the cross-expert boundary reconciliation ablation. Direct concatenation (A0) exhibits pronounced discontinuities across joint displacement, angular change, and velocity jumps. Bilateral translation blending with spherical linear interpolation (A1) substantially alleviates these artifacts by smoothing both sides of the boundary. Our boundary projection mechanism (A2) further achieves the lowest errors across all three metrics, effectively suppressing cross-expert discontinuities while keeping preceding motion segments fixed.

\newpage
\subsection{Synthesized Data Diversity Evaluation}
\label{sec:diversity}

\input{tables/tabS2_diversity}

\looseness=-1 Since World2HOSI is designed as a comprehensive HOSI data synthesis engine, verifying the coverage and diversity of its synthesized outputs is essential. To this end, separate from the main benchmark evaluation, we construct a generated set of 90 sequences spanning more varied tasks. We compare this set against equally sized random subsets of BEHAVE~\citep{bhatnagar22behave}, TRUMANS~\citep{jiang2024scaling}, and LINGO~\citep{jiang2024autonomous}. Our set contains 60 interactions with dynamic objects and 30 static interactions across different scenes. For action-semantic diversity~\citep{kister2026inhabit}, a VLM extracts action and target-object text labels from each sequence. A shared text encoder~\citep{reimers2019sentence} maps these labels to unit-normalized embeddings, whose average cosine distance measures diversity. For pose diversity, we compute the average pairwise distance (APD) over SMPL-X joints in uniformly sampled frames~\citep{kister2026inhabit,SMPL-X:2019}. Table~\ref{tab:diversity} shows that our synthesized data achieve the highest diversity across actions, target objects, and poses, demonstrating World2HOSI's breadth as a data synthesis engine.

\subsection{Additional Qualitative Results on the InfBaGel Benchmark}
\label{app:benchmark2}

Figure~\ref{fig:infbagel2} provides more qualitative comparisons between InfBaGel and World2HOSI on two additional manipulation sequences from the InfBaGel benchmark, specifically involving \texttt{monitor} and \texttt{smalltable}. As highlighted by the red circles across key hand-object contact regions and state transitions, InfBaGel frequently suffers from severe interpenetration and unstable grasping poses when lifting, moving, and placing these objects. In contrast, World2HOSI maintains consistent, physically plausible hand-object contacts and smooth full-body trajectories throughout both sequences, further demonstrating the robustness of our multi-expert harness on diverse object geometries.

\input{figures/figS2_state_dependencies}

\subsection{More results of Long-Horizon Tasks with State-Update Dependencies}
\label{app:state}

To further demonstrate the efficacy of our closed-loop orchestration and state-update mechanism in handling cross-stage state dependencies, we present two representative tasks in unseen scenarios. In the first task, the human must first move an obstacle blocking a doorway before traversing into the adjacent room through the cleared passage. In the second task, the human is required to relocate a chair to a designated open area and subsequently sit on the chair at its new location. Crucially, neither task can be resolved via forward pre-scripted workflows, as subsequent sub-tasks strictly depend on the actual state resulting from preceding interactions. Under our closed-loop formulation, whenever an underlying expert model finishes execution, the system synchronously updates the ISP. This updated environment state is then fed back to determine the initialization conditions, path feasibility, and interaction targets for the subsequent task stage. As shown in Figure~\ref{fig:state}, by updating the shared world state after each stage, World2HOSI enables subsequent motions to be generated based on the updated object positions and scene, thereby supporting multi-stage human–object–scene interactions with explicit state dependencies.

\input{figures/figS3_generation_prompt}

\subsection{Details for GPT-6-Astra Baseline}
\label{app:prompt}
For the standalone GPT-6-Astra~\citep{openai2026gpt6astra} baseline evaluated in the main text, the model is conditioned on task instructions, scene geometry, and initial human and object states to generate executable motion-scripting code that jointly synthesizes human poses and object trajectories. The complete user prompts used for this baseline are provided in Figure~\ref{fig:motionprompt}.

\looseness=-1 \textbf{Qualitative Comparison.} Complementing the quantitative evaluation in the main paper, Figure~\ref{fig:GPT} presents qualitative comparisons between World2HOSI and GPT-6-Astra. Although GPT-6-Astra captures high-level task semantics, its directly generated motions appear noticeably stiff and mechanical. More critically, joint human-object generation is inherently challenging, as a result, GPT-6-Astra fails to guarantee plausible hand-object contact or natural interaction dynamics, frequently exhibiting detached grasps, unnatural object poses (highlighted in red circle). In contrast, World2HOSI maintains plausible contact and produces substantially more natural interactions.

\input{figures/GPT}

\input{tables/tabS3_task_progress}

\subsection{Definitions of Aggregate Metrics for Shared Tasks}
\label{app:metrics}

\subsubsection{Task Progress Score \texorpdfstring{$S_t$}{St}}
\label{app:progress}

For each carrying, pushing, or pulling subtask, we assign weights of 1, 2, 2, and 1 to approach, initiation, movement, and completion, respectively. For sitting or lying down, approach, pose change, and the target pose receive weights of 1, 2, and 2. Observed evidence at each stage is mapped to a partial-completion value in $[0,1]$ and multiplied by the stage weight. The total score is $100\times$ the sum of earned points divided by the sum of available points.

Stage recognition proceeds in temporal order. Once evidence of a stage is observed, evaluation may advance to the next stage even if the current stage has not received full credit. High scores in later stages cannot compensate for entirely missing prerequisite actions.

\subsubsection{Geometry Score \texorpdfstring{$S_g$}{Sg}}
\label{app:geometry}

The geometry score $S_g$ assesses motion plausibility independently of task completion. It combines contact quality, collision behavior, and foot behavior with equal weights. 

Let $\mathrm{OCP}$, $P_f$,
$\mathrm{FS}$, and $O_{\mathrm{feet}}$
denote these means. The contact score is
$\mathrm{OCP}$, and the collision score is
$100-P_f$, with both percentages expressed
on a 0-100 scale. The mean foot sliding
$\mathrm{FS}$ (cm/frame) and mean foot-ground
distance $O_{\mathrm{feet}}$ (cm) are converted
to scores as $100/(1+\mathrm{FS})$ and
$100/(1+O_{\mathrm{feet}}/5)$, respectively.
Their average forms the foot score. The final score is
\begin{equation}
S_g =
\frac{\mathrm{OCP}+100-P_f}{3}
+\frac{1}{6}
\left(
\frac{100}{1+\mathrm{FS}}
+
\frac{100}{1+O_{\mathrm{feet}}/5}
\right).
\end{equation}

\input{figures/figS4_grasp_refinement}

\Needspace{11\baselineskip}
\subsection{Physics-Based Refinement Details}
\label{app:physics-details}
\noindent \textbf{VLM-guided Motion Restoration.} To guide motion restoration, we employ GPT-6-Astra for refinement~\citep{openai2026gpt6astra}. Although prior baseline comparisons show that Astra produces mechanical and stiff full-body motions when generating interactions directly without our harness constraints, its rich commonsense knowledge makes it well suited for locally refining implausible grasps. We provide the prompt we use in Figure~\ref{fig:prompt2}. Taking multi-view hand viewpoints, current hand and object states, and target object geometry as inputs, the prompt instructs Astra to identify plausible grasping regions while keeping body joints and object states unchanged and suppressing unnatural poses or excessive penetration, providing refined references for downstream residual tracking.

\noindent \textbf{Bounded-Residual Control.} For tasks requiring physical execution, the generated kinematic sequence supplies a reference for a bounded-residual controller. Objects move solely through physical human contact, the controller neither replays object trajectories nor applies frame-by-frame external forces to reproduce them. It must establish and maintain contacts that drive object motion while preserving interaction intent. Following the residual-tracking paradigm~\citep{zhao2026real2sim,zhao2025resmimic}, the policy predicts residuals around the reference joint state $q_{\mathrm{ref}}(t)$ rather than generating complete joint targets from scratch:
\begin{equation}
\begin{aligned}
q_{\mathrm{target}}(t)=\operatorname{clip}\!\Bigl(&
q_{\mathrm{ref}}(t)+m\odot b\odot\tanh(a(t)),\\
& q_{\min},q_{\max}\Bigr).
\end{aligned}
\label{eq:app-residual-control}
\end{equation}
Here, $a(t)$ is the policy output, $m$ masks correctable joints, $b$ sets per-joint residual bounds, and $q_{\min},q_{\max}$ specify joint limits. The $\operatorname{clip}$ operator clamps corrected targets to valid joint ranges. Joint proportional-derivative (PD) controllers execute $q_{\mathrm{target}}$. Residual bounds and joint limits restrict the search to a neighborhood of the reference, permitting posture adjustments for contact while preserving the overall motion structure as closely as possible.

\noindent \textbf{Reward Design and Optimization.} The reward combines reference tracking and smoothness terms~\citep{hoifhli}:
\begin{equation}
\begin{aligned}
r_t &= r_t^{\mathrm{track}} + \lambda_s r_t^{\mathrm{acc}},\\
r_t^{\mathrm{track}}
&= \lambda_{\mathrm{ori}} r_t^{\mathrm{ori}}
+ \lambda_{\mathrm{pos}} r_t^{\mathrm{pos}}
+ \lambda_{\mathrm{hand}} r_t^{\mathrm{hand}} .
\end{aligned}
\label{eq:app-reward}
\end{equation}
where
$(\lambda_{\mathrm{ori}},\lambda_{\mathrm{pos}},
\lambda_{\mathrm{hand}},\lambda_s)=(0.4,0.4,0.2,0.05)$. The orientation term $r_t^{\mathrm{ori}}$ measures disagreement between simulated and reference human or object orientations, $r_t^{\mathrm{pos}}$ measures their positional disagreement. The hand-relative term $r_t^{\mathrm{hand}}$ measures disagreement in hand-object geometry in the object's local frame. The hand-relative term specifies their geometric relationship without explicitly requiring contact, avoiding the imposition of reference penetration or floating artifacts as execution targets. The term $r_t^{\mathrm{acc}}$ penalizes excessive acceleration to reduce high-frequency jitter. Using proximal policy optimization (PPO)~\citep{schulman2017proximal}, we search for a physically executable solution near the reference rather than generate a new interaction. This preserves its overall structure while reducing unintended floating, penetration, and jitter, enabling contact-driven object motion with improved physical executability.

\noindent \looseness=-1 \textbf{Implementation.} We perform this physical refinement in Isaac Gym. The policy observations comprise the current poses and velocities of the human and manipulated object, together with their reference poses. All experiments are conducted on a single A800 GPU, where we optimize a separate policy for each reference sequence using 4,096 parallel environments. Across the evaluated sequences, the mean PPO training time is approximately 88.6 minutes per sequence, depending on the duration of the original sequence: training takes around 40 minutes on the shortest sequence (approximately 6.3 seconds), whereas the longest long-horizon motions (approximately 36 seconds) require around 2 hours.

\subsection{User Study}
\label{app:user-study}

\looseness=-1 The questionnaire contains eight tasks. For each task, World2HOSI, InfBaGel, and HOSIG each generate one kinematic video, yielding 24 videos in total. The tasks correspond to T01, T02, T03, T04, T05, T06, T11, and T12 in the evaluated version. Each group of three videos uses identical human appearance, scene conditions, camera settings, and rendering parameters. The videos are randomly ordered and displayed side by side with the labels A, B, and C, without revealing the method names.

Each task has two independent questions: select the video with the best task completion and select the one with the best interaction quality. The response options are methods A, B, and C, and \emph{No Preference}.

With the informed consent of the participants, all 30 participants had answered both questions for all eight tasks. Each evaluation criterion therefore has 240 responses.

%% file: figures/figS1_infbagel.tex
\begin{figure*}[!b]
\centering
\includegraphics[width=\linewidth]{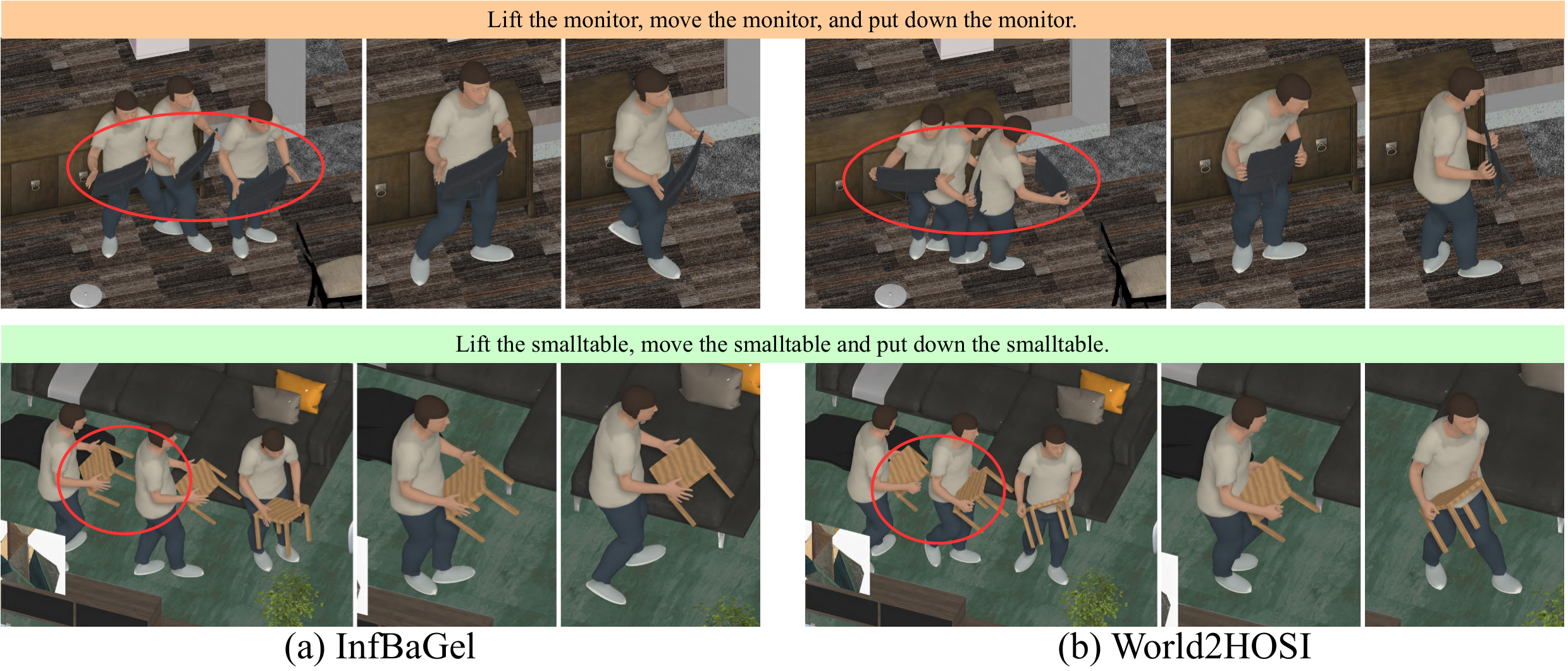}
\caption{Qualitative comparison between (a) InfBaGel and (b) World2HOSI. Red circles highlight key hand–object contact regions.}
\label{fig:infbagel2}
\end{figure*}

%% file: tables/tabS1_reconciliation.tex
\begin{table}[t]
\centering
\caption{\raggedright Ablation of cross-expert reconciliation. We compare different boundary treatments to assess the improvement in motion smoothness from cross-expert reconciliation.}
\label{tab:reconciliation}
\begingroup
\small
\setlength{\tabcolsep}{3pt}
\renewcommand{\arraystretch}{1.15}
\begin{tabular*}{\linewidth}{@{\extracolsep{\fill}}lccc@{}}
\toprule
Metric $\downarrow$ & Concat. (A0) & Bi-Blend (A1) & Proj. (A2) \\
\midrule
Joint Disp. (cm) & 1.135 & 0.840 & \textbf{0.593} \\
Joint Angle ($^{\circ}$) & 4.615 & 1.224 & \textbf{0.397} \\
Velocity Jump (cm/s) & 33.437 & 0.637 & \textbf{0.208} \\
\bottomrule
\end{tabular*}
\endgroup
\end{table}

%% file: tables/tabS2_diversity.tex
\begin{table}[t]
\centering
\caption{\raggedright Comparison of motion and object diversity across datasets. We construct a small-scale generated set and compare it with equally sized, randomly sampled subsets of external datasets.}
\label{tab:diversity}

\begingroup
\small
\setlength{\tabcolsep}{4pt}
\renewcommand{\arraystretch}{1.15}

\begin{tabular*}{\linewidth}{@{\extracolsep{\fill}}lccc@{}}
\toprule
Dataset & Action Div. $\uparrow$ & Target Obj. Div. $\uparrow$ & APD (cm) $\uparrow$ \\
\midrule
BEHAVE & 0.5599 & 0.6481 & 17.72 \\
TRUMANS & 0.6147 & 0.6534 & 18.70 \\
LINGO & 0.5267 & 0.6774 & 13.56 \\
World2HOSI & \textbf{0.6284} & \textbf{0.6796} & \textbf{23.57} \\
\bottomrule
\end{tabular*}
\endgroup
\end{table}

%% file: figures/figS2_state_dependencies.tex
\begin{figure*}[!b]
\centering
\includegraphics[width=\linewidth]{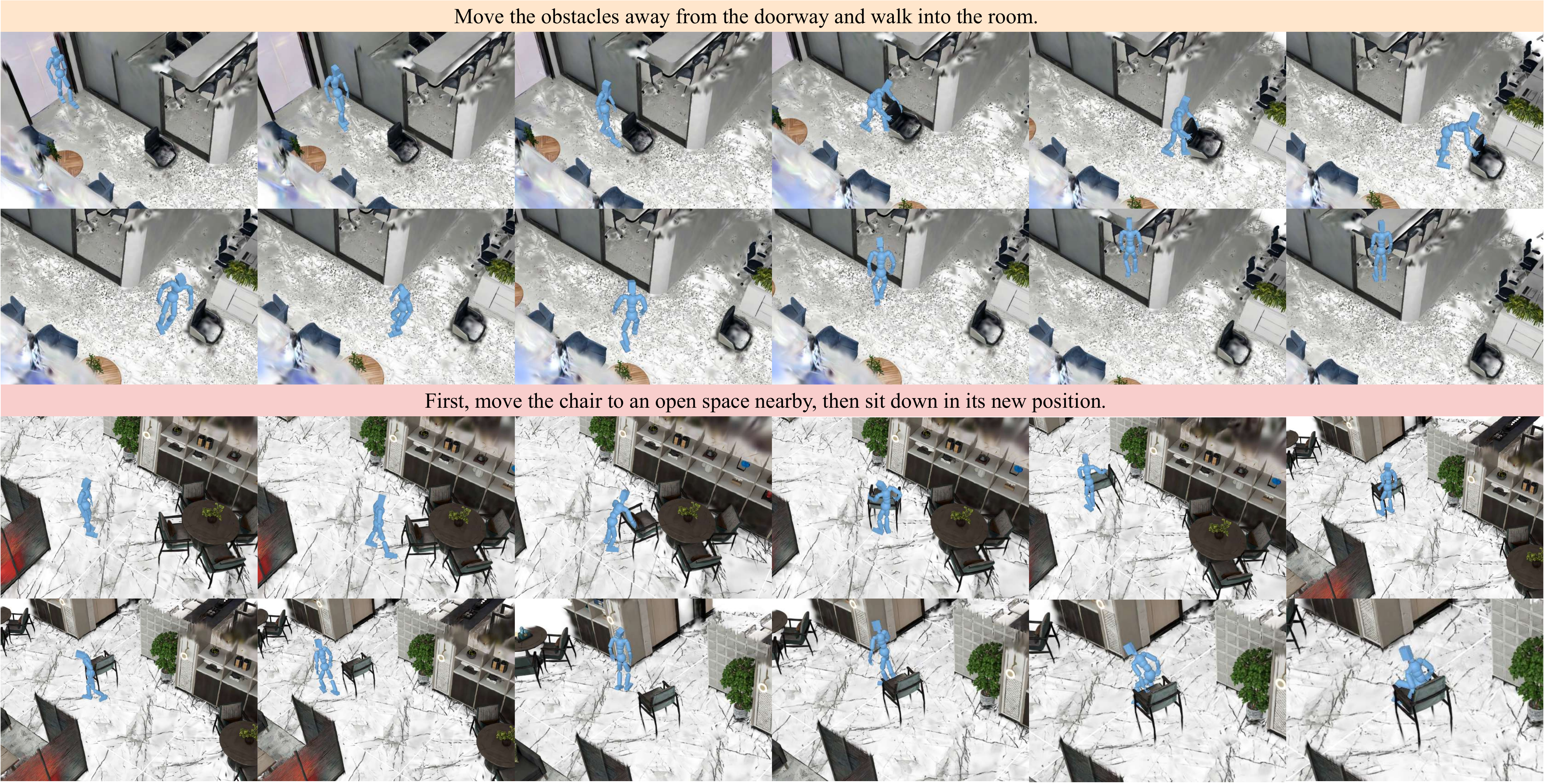}
\caption{Long-horizon interaction tasks with state-update dependencies. Top: Removing an obstacle blocking a doorway before entering the room. Bottom: Moving a chair to a new location before sitting on it. Both examples illustrate how World2HOSI uses scene states updated by preceding interactions to provide consistent spatial conditions and interaction constraints for subsequent motion generation.}
\label{fig:state}
\end{figure*}

%% file: figures/figS3_generation_prompt.tex
\begin{figure*}[t]
\centering
\includegraphics[width=\linewidth]{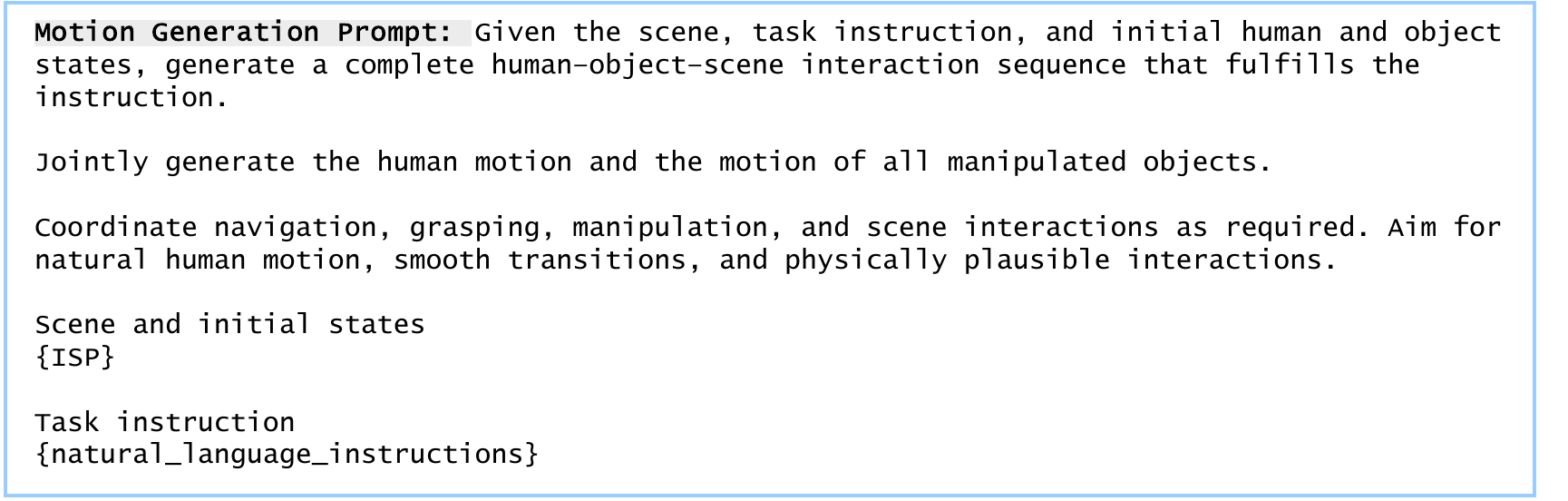}

\caption{Prompts for GPT-6 to generate human motions. }
\label{fig:motionprompt}
\end{figure*}

%% file: figures/GPT.tex
\begin{figure*}[t]
\centering
\includegraphics[width=\linewidth]{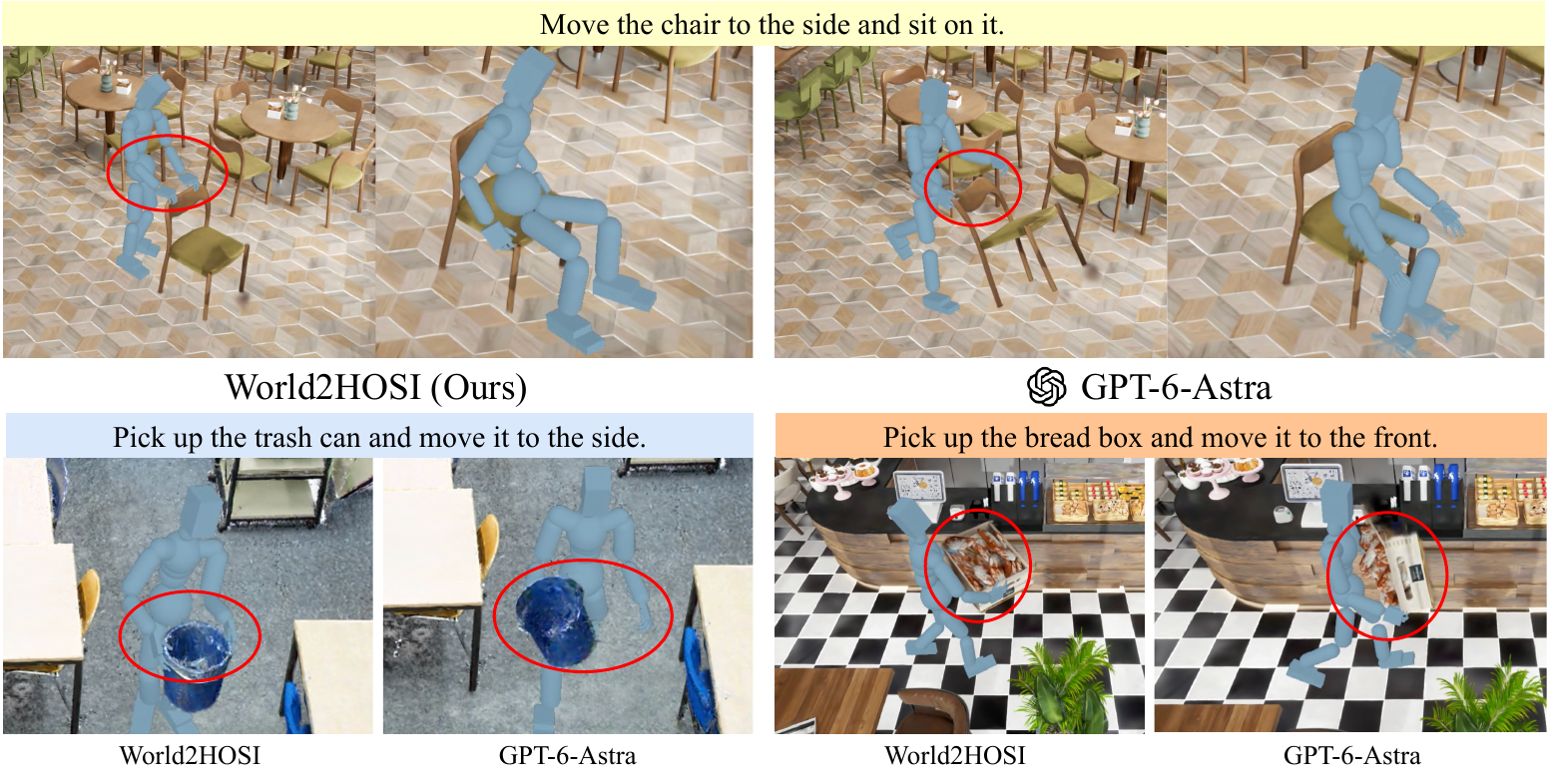}
\caption{Qualitative comparison between World2HOSI and GPT-6-Astra, with red circles highlighting key interaction regions. }
\label{fig:GPT}
\end{figure*}

%% file: tables/tabS3_task_progress.tex
\begin{table*}[!b]
\centering
\caption{Evidence and scoring rules for the task progress score $S_t$.}
\label{tab:task-progress-rules}
\begingroup
\small
\renewcommand{\arraystretch}{1.15}
\begin{tabularx}{\linewidth}{@{}p{0.27\linewidth}X@{}}
\toprule
Evidence & Scoring rule \\
\midrule
Human-target distance & Full credit when the distance is at most 10\,cm, zero at or beyond 60\,cm, and linear interpolation between these thresholds. \\
Approach / arrival distance & A target is considered approached within 50\,cm and reached within 30\,cm. \\
Initiation / full-credit displacement & Manipulation is considered initiated when the object has moved 1\,cm; the initiation component receives full credit at 3\,cm. \\
Motion speed / stopping observation & The object is considered in motion at speeds of at least 0.02\,m/s. Placement or stopping is recognized when its position varies by no more than 5\,cm over a continuous 0.2\,s interval. \\
Object target region & Full credit when the final object position is within 0.50\,m of the target, and zero when it is more than 1.00\,m away. \\
Sitting pose change & The full-credit scales are a 0.15\,m decrease in pelvis height and $60^{\circ}$ of knee flexion. \\
Sitting / lying evidence & A VLM inspects the final frame to determine whether the human has reached the target pose. \\
\bottomrule
\end{tabularx}
\endgroup
\end{table*}

%% file: figures/figS4_grasp_refinement.tex
\begin{figure*}[t]
\centering
\includegraphics[width=\linewidth]{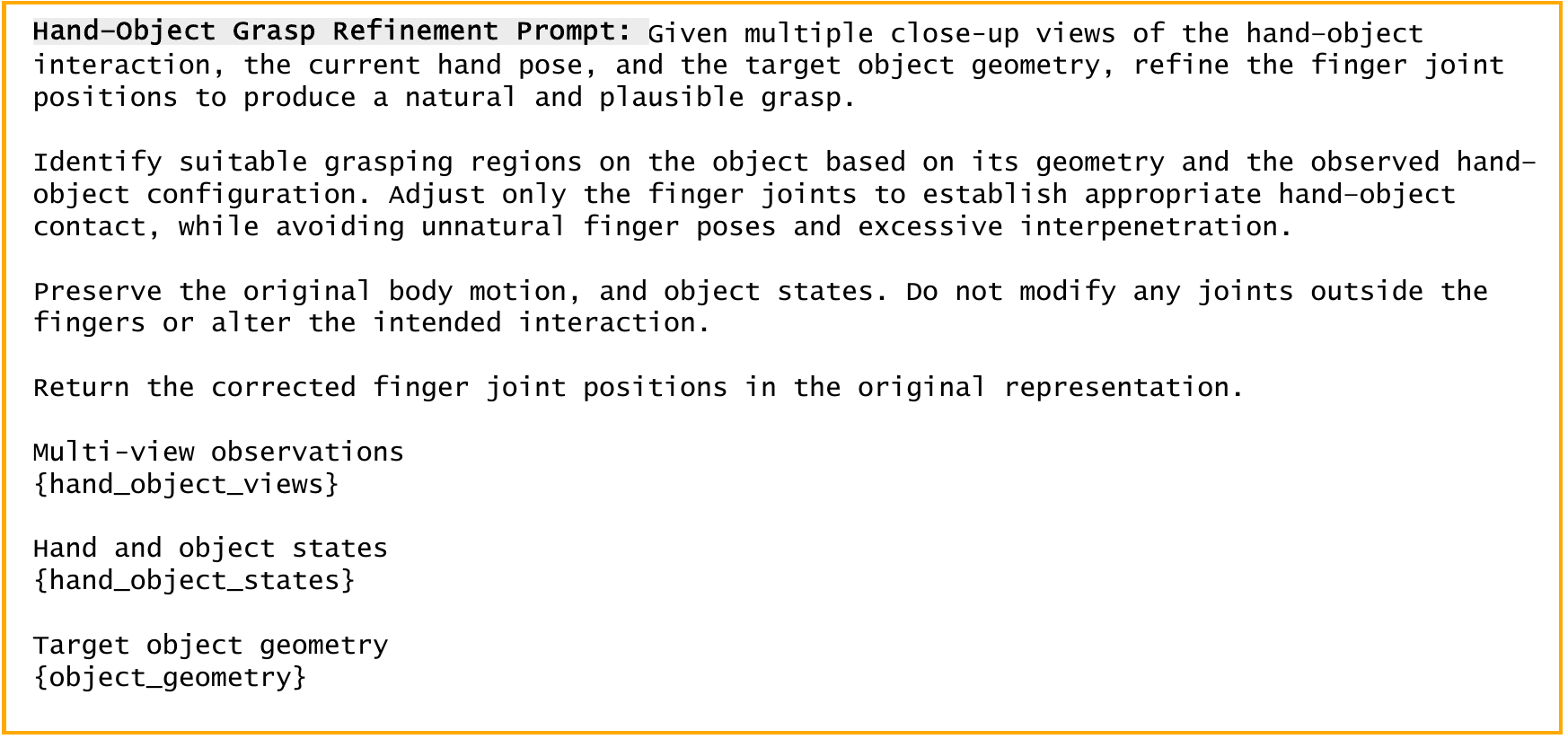}

\caption{Prompts for VLM to refine motions. }
\label{fig:prompt2}
\end{figure*}